\documentclass{article} 
\usepackage[final]{neurips_2025} 

\usepackage[utf8]{inputenc} 
\usepackage[T1]{fontenc}    
\usepackage{hyperref} 
\usepackage{subcaption}  
\hypersetup{
 colorlinks=true,
 linkcolor=red,
 citecolor=cyan,
 filecolor=magenta, 
 urlcolor=magenta,
 }
\usepackage{url}            
\usepackage{booktabs}       
\usepackage{amsfonts}       
\usepackage{nicefrac}       
\usepackage{microtype}      
\usepackage{xcolor}         
\usepackage{amsmath}     
\usepackage{xspace}       
\usepackage{enumitem}   
\usepackage{textcomp}
\usepackage{stfloats}
\usepackage{verbatim}
\usepackage{graphicx}
\usepackage{subcaption}
\usepackage{amssymb}
\usepackage{cite}

\def\ie{\emph{i.e., }}
\def\etc{\emph{etc.}}
\def\eg{\emph{e.g., }}
\usepackage{multicol,multirow} 
\usepackage{threeparttable}

\title{Matrix-Game: Interactive World Foundation Model}

\author{%
  \textbf{Yifan Zhang}\thanks{Equal contribution.}\quad
  \textbf{Chunli Peng}$^{*}$\quad
  \textbf{Boyang Wang}$^{*}$\thanks{Project Lead.}\quad
  \textbf{Puyi Wang}\quad 
  \textbf{Qingcheng Zhu} \\
  \textbf{Fei Kang}\quad
  \textbf{Biao Jiang}\quad
  \textbf{Zedong Gao}\quad
  \textbf{Eric Li}\quad
  \textbf{Yang Liu}\quad
  \textbf{Yahui Zhou} \\
  \\
  Skywork AI \\
  Project page: \url{https://matrix-game-homepage.github.io}
}

\begin{document}
\maketitle

\begin{abstract}
We introduce Matrix-Game, an interactive world foundation model for controllable game world generation. Matrix-Game is trained using a two-stage pipeline that first performs large-scale unlabeled pretraining for environment understanding, followed by action-labeled training for interactive video generation. To support this, we curate Matrix-Game-MC, a comprehensive Minecraft dataset comprising over 2,700 hours of unlabeled gameplay video clips and over 1,000 hours of high-quality labeled clips with fine-grained keyboard and mouse action annotations. Our model adopts a controllable image-to-world generation paradigm, conditioned on a reference image, motion context, and user actions. With over 17 billion parameters, Matrix-Game enables precise control over character actions and camera movements, while maintaining high visual quality and temporal coherence. To evaluate performance, we develop  GameWorld Score, a unified benchmark measuring visual quality, temporal quality, action controllability, and physical rule understanding for Minecraft world generation. Extensive experiments show that Matrix-Game consistently outperforms prior open-source Minecraft world models—including Oasis and MineWorld—across all metrics, with particularly strong gains in controllability and physical consistency. Double-blind human evaluations further confirm the superiority of Matrix-Game, highlighting  its ability to generate perceptually realistic and precisely controllable videos across diverse game scenarios. To facilitate future research on interactive image-to-world generation, we will open-source the Matrix-Game model weights and the GameWorld Score benchmark at \url{https://github.com/SkyworkAI/Matrix-Game}.

\end{abstract}

\begin{figure}[p]
\centering
\begin{subfigure}{0.99\linewidth}
    \centering
    \includegraphics[width=\linewidth]{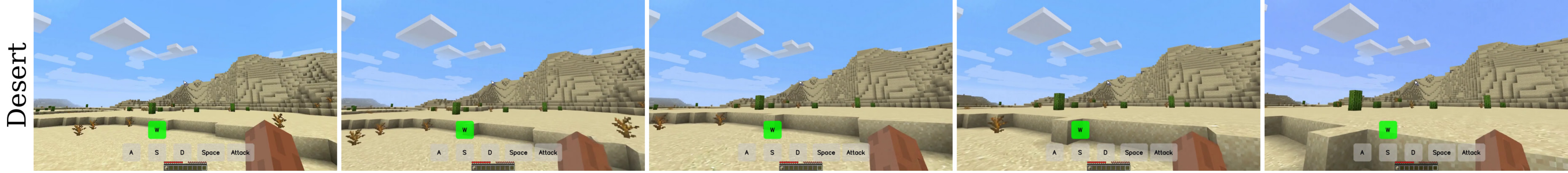}
    \caption{Desert}
    \vspace{0.1in}
\end{subfigure}
\begin{subfigure}{0.99\linewidth}
    \centering
    \includegraphics[width=\linewidth]{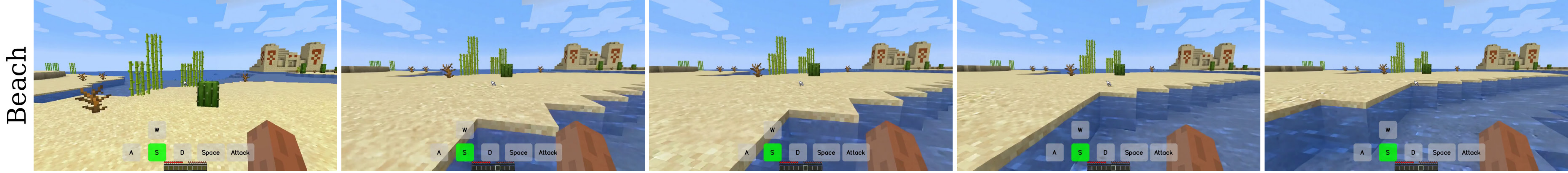}
    \caption{Beach}
    \vspace{0.1in}
\end{subfigure}  

\begin{subfigure}{0.99\linewidth}
    \centering
    \includegraphics[width=\linewidth]{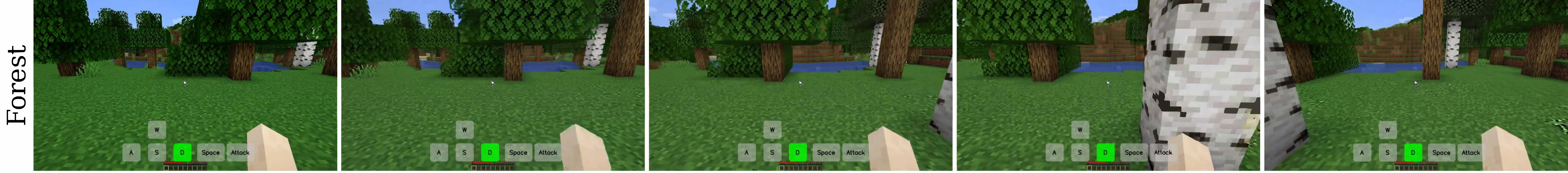}
    \caption{Forest}
    \vspace{0.1in}
\end{subfigure}  

\begin{subfigure}{0.99\linewidth}
    \centering
    \includegraphics[width=\linewidth]{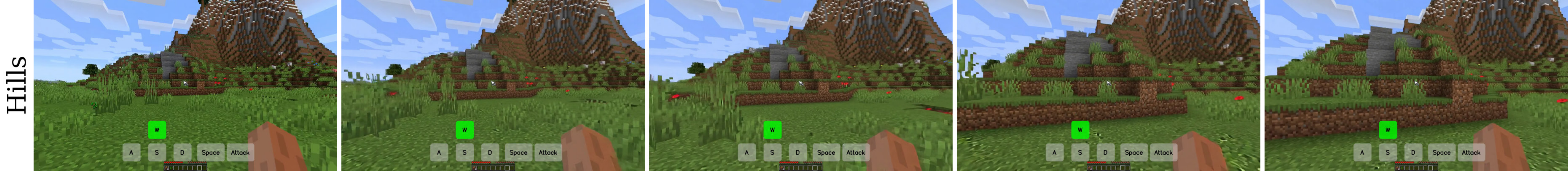}
    \caption{Hills}
    \vspace{0.1in}
\end{subfigure}  

\begin{subfigure}{0.99\linewidth}
    \centering
    \includegraphics[width=\linewidth]{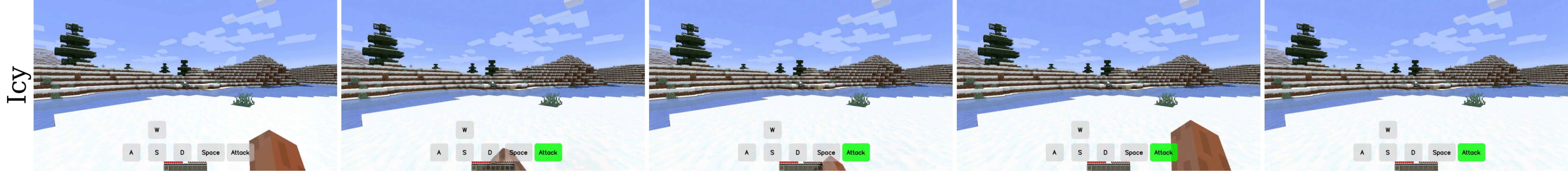}
    \caption{Icy}
    \vspace{0.1in}
\end{subfigure}  

\begin{subfigure}{0.99\linewidth}
    \centering
    \includegraphics[width=\linewidth]{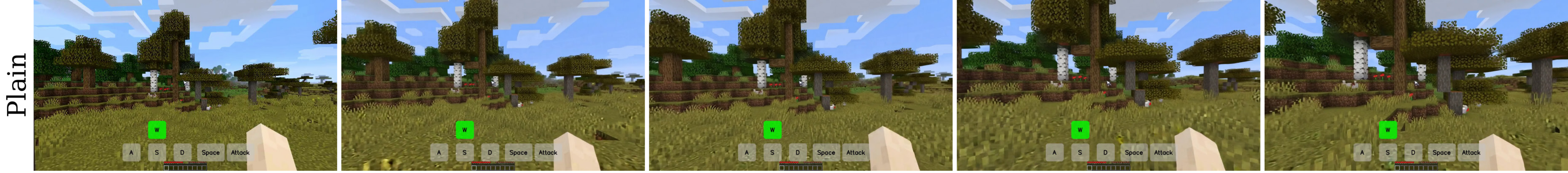}
    \caption{Plain}
    \vspace{0.1in}
\end{subfigure}  

\begin{subfigure}{0.99\linewidth}
    \centering
    \includegraphics[width=\linewidth]{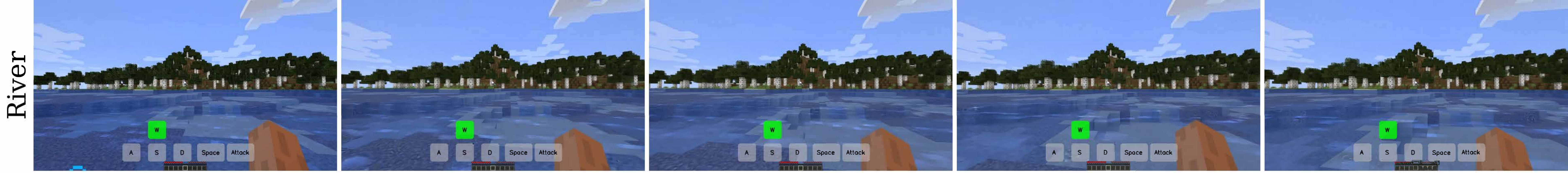}
    \caption{River}
    \vspace{0.1in}
\end{subfigure}  

\begin{subfigure}{0.99\linewidth}
    \centering
    \includegraphics[width=\linewidth]{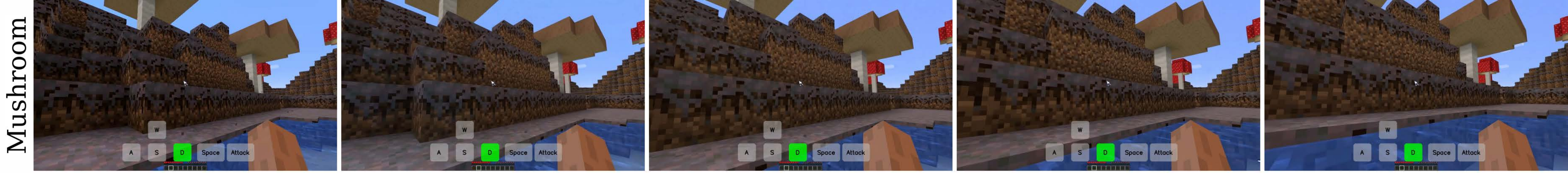}
    \caption{Mushroom} 
\end{subfigure}  
\caption{Controllable world generation results of \textit{Matrix-Game} across distinct Minecraft scenarios. These demos illustrate the model's ability to handle diverse environments, ranging from \textit{desert}, \textit{beach}, and \textit{forest} to more challenging settings like \textit{mushroom} and \textit{icy} biomes, while accurately responding to user control signals.}
\label{fig:demo_scenarios}
\end{figure}

\begin{figure}[t]
    \vspace{-0.2in}
    \centering 
    \includegraphics[width=0.68\linewidth]{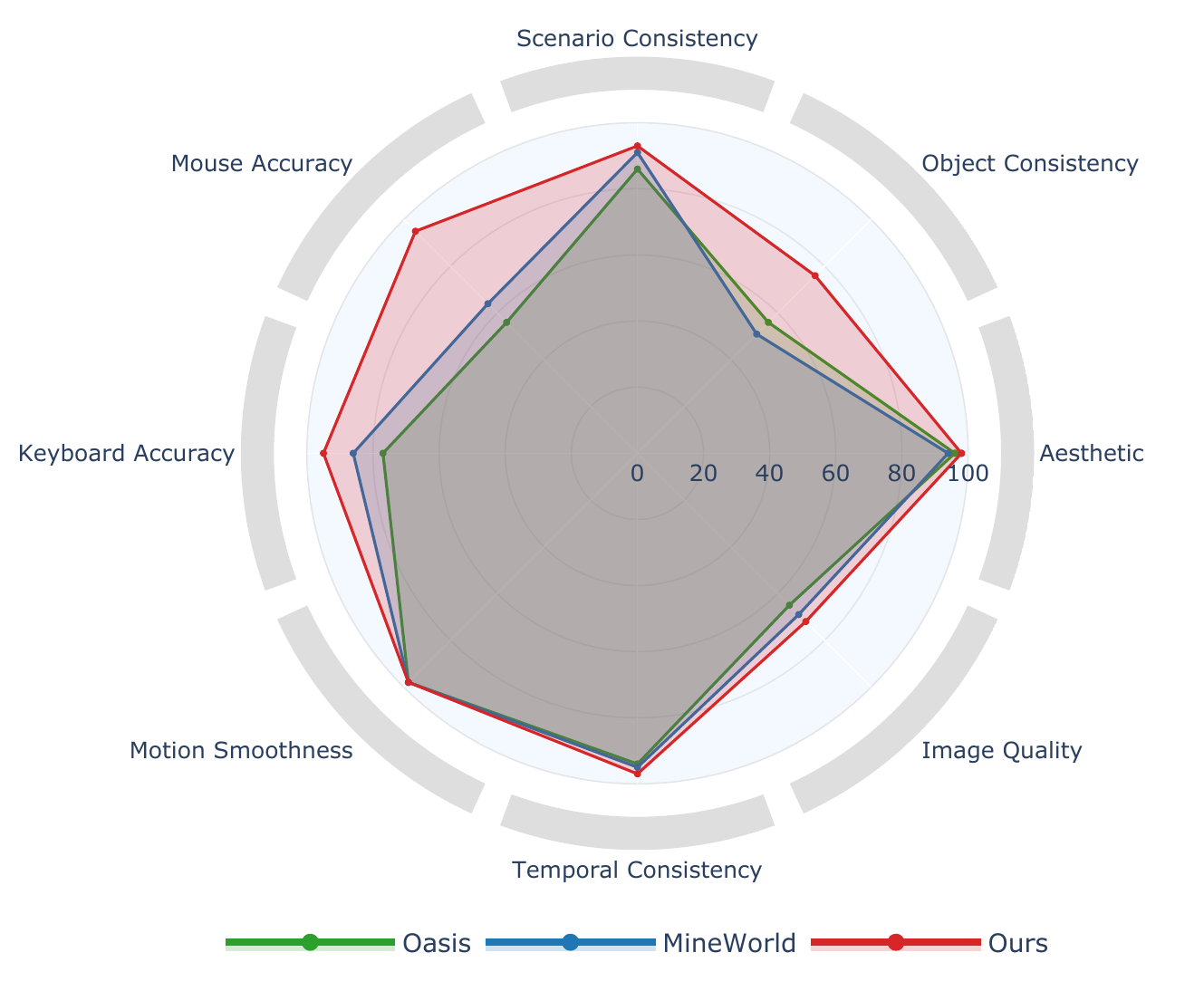}
    \vspace{-0.1in}
    \caption{Model performance under our {GameWorld Score} benchmark, covering 8 key dimensions: Image Quality, Aesthetic (scaled $\times$2 for visualization), Temporal Consistency, Motion Smoothness, Keyboard Accuracy, Mouse Accuracy, Object Consistency and Scenario Consistency.  Our method outperforms Oasis~\cite{oasis2024} and MineWorld~\cite{guo2025mineworld} in all aspects, particularly in controllability (keyboard and mouse accuracy) and physical consistency, while maintaining high visual and temporal quality.
}\vspace{-0.2in}
    \label{fig:radar_1}
\end{figure}

\section{Introduction}

World models~\cite{WorldModels,zhu2024sora} are fundamental to intelligent agents, enabling them to perceive, simulate, and reason about the dynamics of their environments. By internalizing the structure and behavior of the external world, these models support a wide spectrum of downstream tasks, including autonomous driving~\cite{gao2024vista,wang2024drivedreamer}, embodied intelligence~\cite{gupta2024essential,hafner2024embodied}, and generative game engines~\cite{guo2025mineworld,alonso2024diffusion}. Effective world models empower agents not only to interpret observations but also to anticipate outcomes and plan actions in open-ended, uncertain environments~\cite{liu2024world,kim2024openvla,ma2024survey}.

In recent years, video diffusion models have emerged as a leading paradigm for world modeling~\cite{openai2024worldsim,parkerholder2024genie2,agarwal2025cosmos,yang2024position,yanglearning}. Their ability to learn fine-grained spatial-temporal dynamics and generate visually coherent videos makes them particularly suited for simulating evolving worlds. Moreover, when conditioned on structured inputs such as keyboard actions and camera movements, video diffusion models provide a flexible and scalable approach to modeling complex agent-environment interactions with high fidelity~\cite{feng2024matrix,parkerholder2024genie2,guo2025mineworld}.

While video-based world models present a promising direction for simulating and understanding dynamic worlds, they face several significant challenges. First, acquiring high-quality training data is non-trivial. Interactive video datasets with rich annotations (e.g., precise actions, camera movement) are scarce and expensive to collect, especially at scale. Second, modeling the physical dynamics of the world and achieving fine-grained controllability over time remains difficult. World models need not only to generate visually coherent videos but also to respond accurately to structured control signals. Third, the lack of a standardized evaluation benchmark makes it hard to objectively compare models or quantify improvements, limiting the research in this area. 

To address the above challenges, we introduce Matrix-Game, an interactive world foundation model designed for game world generation. Our approach comprises three core components. First, we construct Matrix-Game-MC, a large-scale dataset specifically tailored for Minecraft world modeling, comprising both unlabeled gameplay video clips and richly annotated action-labeled video data. This dataset enables the learning of complex environment dynamics and interaction patterns at scale. Second, we develop Matrix-Game, a diffusion-based image-to-world generation model that supports interactive video generation conditioned on user inputs such as keyboard commands and mouse-driven camera movements. This architecture emphasizes controllability, temporal coherence, and visual fidelity in dynamic world environments. Third, to systematically evaluate world models, we develop GameWorld Score, a unified benchmark encompassing multiple evaluation dimensions (including visual quality, temporal quality, controllability, and physical rule understanding), offering a comprehensive framework for quantitative assessment of Minecraft world models.

To train Matrix-Game, we develop a two-stage strategy: an initial stage of unlabeled training for game world understanding, followed by action-labeled training for interactive world generation. Building on this training pipeline, we scale up both the model capacity (to 17B parameters) and the training data. As a result, Matrix-Game is able to generate visually coherent scenes that reflect realistic motion patterns, respond accurately to user controls (including keyboard movement and mouse-driven camera actions) and maintain high temporal consistency throughout the video. As shown in Figure~\ref{fig:radar_1} and Table~\ref{table:exp_result}, it consistently outperforms leading open-source Minecraft world models such as Oasis~\cite{oasis2024} and MineWorld~\cite{guo2025mineworld}, with particularly strong gains in action controllability and physical rule understanding. Furthermore, as illustrated in Figure~\ref{fig:demo_scenarios}, Matrix-Game produces high-quality, controllable videos that align with in-game physics across a variety of Minecraft scenarios. By integrating visual perception with fine-grained user control, Matrix-Game redefines video generation as an interactive process of exploration and creation, empowering users to observe, guide, and construct coherent virtual worlds from a single reference image.
 
\section{Related Work}
\subsection{Video Diffusion Models}
With the rapid advancement of diffusion models~\cite{sohl2015deep,ho2020denoising,songscore}, visual content generation has witnessed remarkable progress across both image~\cite{rombach2022high,podell2023sdxl,zhang2023hipa,zhang2023expanding,dai2024one,lin2025drivegen} and video domains~\cite{tian2024emo,zheng2024memo,blattmann2023stable,lin2024open,yang2024cogvideox}. Notably, video generation has evolved beyond traditional U-Net-based backbones, transitioning toward Transformer-based architectures~\cite{peebles2023scalable,vaswani2017attention} that offer enhanced scalability and temporal modeling capabilities. This shift has empowered modern video diffusion models to synthesize high-quality, temporally coherent videos of significantly longer durations~\cite{openai2024worldsim}.

The impressive fidelity has sparked growing interest in using video diffusion models to simulate complex world environments by implicitly learning physical laws, object dynamics, and causal interactions from raw video data~\cite{openai2024worldsim,parkerholder2024genie2,agarwal2025cosmos,yang2024position,yanglearning}. As a result, video diffusion models are increasingly viewed as promising world models for embodied agents and interactive systems, where reasoning about physical causality, spatial continuity, and controllable dynamics is essential for tasks such as planning, navigation, and decision-making.

\subsection{Controllable Video Generation}
While text descriptions serve as the primary input for text-to-video generation~\cite{kong2024hunyuanvideo,wang2025wan}, they often lack sufficient precision to capture complex spatial-temporal details, leading to ambiguous or underspecified outputs. To overcome this limitation, recent research has introduced additional control modalities to better align generation with user intent and improve content determinism. Among these, reference image guidance has proven effective for enhancing both visual fidelity and temporal consistency by anchoring the generated video to a concrete visual context~\cite{blattmann2023stable,ni2023conditional}.

Beyond those static signals as conditions, Direct-a-Video~\cite{yang2024direct} introduces a camera embedder to influence perspective during generation, but its capabilities are constrained to only three coarse attributes (e.g., yaw/pitch/zoom). CameraCtrl~\cite{he2024cameractrl} and MotionCtrl~\cite{wang2024motionctrl} go further by introducing fine-grained trajectory-level camera guidance, offering temporally-continuous control over position and orientation. These methods enable dynamic scene exploration and user-driven cinematic effects. In parallel, World-Model-based methods~\cite{hu2023gaia,xiao2025worldmem} emphasize action-conditioned rollout to model physical dynamics. However, most of these approaches are limited by low visual quality or simplified action spaces. In our work, we focus on interactive image-to-world generation, where generation is guided by keyboard and mouse movement actions, enabling precise and intuitive control in open-ended environments.

\subsection{Game Video Generation}
Given the strong modeling capabilities of video diffusion models, an increasing number of studies have begun exploring their applications in game world generation~\cite{alonso2024diffusion,che2024gamegen,bruce2024genie,oasis2024,parkerholder2024genie2,feng2024matrix,yu2025gamefactory,parkerholder2024genie2,xiao2025worldmem,valevski2024diffusion,yang2024playable,guo2025mineworld,yu2025position}. Genie~\cite{bruce2024genie} proposes a foundation model for playable environments built upon video generation. Other methods such as DIAMOND~\cite{alonso2024diffusion}, GameNGen~\cite{valevski2024diffusion}, OASIS~\cite{oasis2024}, and PlayGen~\cite{yang2024playable} leverage diffusion-based generative models to simulate game worlds.
GameGenX~\cite{che2024gamegen} further introduces OGameData, a dataset designed to enable game video generation and controllability. Despite their contributions, many of these approaches, including concurrent works such as GameGenX~\cite{che2024gamegen}, OASIS~\cite{oasis2024}, and WorldMem~\cite{xiao2025worldmem}, tend to overfit to specific game datasets, exhibiting limited generalization across diverse game scenarios and interaction dynamics.

Some recent research works, such as Matrix~\cite{feng2024matrix}, Genie 2~\cite{parkerholder2024genie2}, GameFactory~\cite{yu2025gamefactory}, and MineWorld~\cite{guo2025mineworld}, have attempted to improve control generalization in virtual environments. However, these approaches remain constrained by relatively limited model capacity and dataset scale, making it difficult to effectively capture physical rules and enable accurate interactions across diverse virtual game worlds. In this work, we aim to address these limitations by scaling up both the model size and training data, aiming to advance controllable game world generation with stronger generalization and physical rule understanding. Through this effort, we aim to push the frontier of interactive image-to-world generation, enabling more reliable and versatile agent-environment simulation grounded in user interaction.

\clearpage
\section{Matrix-Game-MC: A Large-scale Dataset for Game World Generation}

Our goal is to develop a world foundation model that can internalize physical dynamics, semantic structures, and support interactive video generation. To achieve this, large-scale, high-quality data is indispensable. We adopt Minecraft as the primary environment due to its diverse biomes, rich agent-environment interactions, and open-ended gameplay, which make it well-suited for learning world modeling. However, acquiring action-labeled Minecraft data through manual gameplay is both time-consuming and resource-intensive. To address this, we supplement training with a large volume of unlabeled gameplay videos to help the model learn motion dynamics and environmental rules. In parallel, we construct an automated pipeline for generating fine-grained, action-labeled video clips in a scalable manner, enabling controllable model training across varied scenarios.

\subsection{Unlabeled Data Collection} 

\textbf{Unlabeled data acquisition protocol.}
The unlabeled training dataset was systematically collected from video resources provided in the MineDojo Dataset~\cite{fan2022minedojo}. We retrieved approximately 6,000 hours of raw gameplay footage through the dataset's official video repositories, which included tutorial content demonstrating core game mechanics, unstructured gameplay recordings, and demonstrations of environmental interactions.

\begin{figure}[t] 
    \centering
    \includegraphics[width=0.99\textwidth]{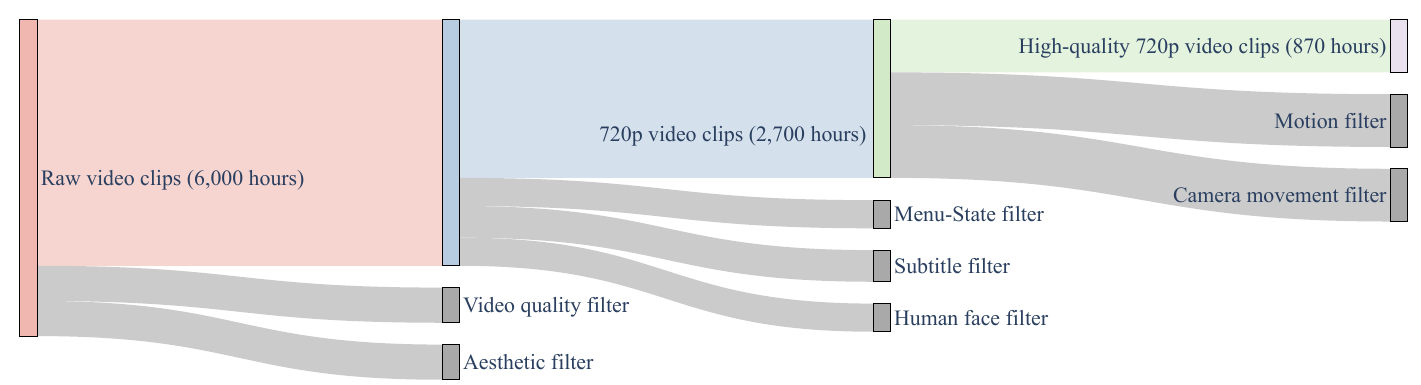}
    \caption{We construct our high-quality unlabeled training data from raw gameplay videos via a three-stage hierarchical filtering pipeline. } 
    \label{fig:datafilter} 
\end{figure}

This diverse collection spans multiple biomes, including forest, desert, and snow terrain ecosystems, offering broad visual and physical coverage of the Minecraft environment. To prepare the data for training, we first employ TransNet V2~\cite{soucek2024transnet} to detect scene transitions and segment raw gameplay videos into single-shot clips. Segmentation is performed using FFmpeg~\cite{ffmpeg} at the detected transition boundaries. Prior to processing, all video data is converted to the \textit{libx264} encoding format for compatibility and efficiency. To mitigate artifacts from gradual transitions or unstable camera motion, we discard the first and last four frames of each segmented clip.
We then introduce a hierarchical filtering framework (illustrated in Figure~\ref{fig:datafilter}) designed to curate high-quality, informative clips from raw gameplay footage. This framework integrates multiple filter modules, each capturing a different aspect of video quality, content, or temporal structure.

\textbf{Unlabeled data filtering pipeline.}
As shown in Figure~\ref{fig:datafilter}, our data filtering pipeline consists of three sequential stages. The first stage focuses on \textit{video quality filtering} and \textit{aesthetic filtering}. The second stage applies \textit{menu-state filtering}, \textit{subtitle filtering}, and \textit{face filtering} to remove non-informative or distracting content. The final stage involves \textit{motion analysis} and \textit{camera movement filtering} to ensure dynamic yet visually stable clips suitable for model training. We provide detailed descriptions of each filter below.

\textit{Video quality filtering.}
We use DOVER~\cite{wu2023dover} to assess video quality, applying genre-specific thresholds to accommodate stylistic diversity across game types. This ensures the retention of videos with sufficient resolution, clarity, and coherence for reliable model training.

\textit{Aesthetic filtering.}
We compute aesthetic scores using the LAION predictor~\cite{LAIONaes}, averaging scores across sampled frames per clip. To account for stylistic diversity across game genres, we apply adaptive, genre-aware thresholds. This ensures selected videos maintain visually coherent and appealing composition, supporting realistic generation.

\textit{Menu-State filtering.}
We use an Inverse Dynamics Model (IDM)~\cite{baker2022video} to detect frames without player input (such as menus, idle states, or loading screens) and exclude them. This ensures the dataset focuses on active gameplay, enhancing the model's ability to learn action-conditioned dynamics and controllable temporal transitions.

\textit{Subtitle filtering}: We apply the CRAFT text detector~\cite{baek2019character} to identify and remove videos with intrusive subtitles, stream banners, or watermarks. By focusing detection on lower-screen and high-risk regions, we preserve in-game text while excluding distracting post-production overlays, ensuring clean visual inputs for training.

\textit{Human face filtering.}
To ensure dataset focus on in-game environments, we use DeepFace~\cite{serengil2024lightface} to detect and filter videos containing streamer face cams or human overlays. By checking for recurring faces in common webcam regions across multiple frames, we eliminate non-game human content, preserving scene purity and preventing the model from learning spurious visual cues.

\textit{Motion filtering}:
To ensure meaningful temporal dynamics, we apply motion filtering using GMFlow~\cite{xu2022gmflow} to compute average optical flow magnitude per clip. Videos with too little motion (\eg static screens) or excessive motion (\eg rapid spinning or scene glitches) are discarded. This bidirectional filtering retains motion-balanced sequences, supporting stable training and improving the model’s ability to learn temporally consistent and controllable video generation.

\textit{Camera movement filtering}: To remove clips with overly aggressive viewpoint changes, we apply camera motion filtering based on angular changes estimated by the Inverse Dynamics Model (IDM)~\cite{baker2022video}. Videos with excessive yaw or pitch rotation (often caused by abrupt mouse movements) are discarded. This filtering step promotes stable and coherent viewpoint trajectories, helping the model learn consistent scene geometry and spatial alignment over time.

\subsection{Labeled Data Creation}

\textbf{Hybrid labeled dataset construction via exploration and simulation.} 
To enable controllable video generation, we construct labeled datasets using two complementary strategies: \textit{Exploration Agent} trajectories derived from gameplay in the MineRL~\cite{minerl2020} environment, and \textit{Unreal Procedural Simulation}, which offers high-fidelity, scriptable environments with precise control annotations.

\textit{Exploration agent.}
We extend the MineRL platform by deploying curriculum-guided VPT agents~\cite{baker2022video}, which are capable of performing long-horizon tasks within the Minecraft world. These agents autonomously explore diverse in-game scenarios, generating a wide range of behavior patterns. We extract per-frame keyboard and mouse actions from these trajectories to construct an action-labeled dataset, sampled at 16Hz. This large-scale dataset provides dense supervisory signals for training control-aware generation models.

\textit{Unreal procedural simulation.}
To complement exploration data with highly structured demonstrations, we build custom environments in Unreal Engine that span various biomes, including urban, desert, and forest settings. Each environment is programmatically designed and instrumented to provide detailed supervision at every frame. Specifically, we collect: (1) Discrete action labels (e.g., movement keys and jump) and continuous gaze vectors (camera pitch/yaw).
(2) Ground-truth kinematic information, including the agent’s position, velocity, and orientation.
(3) Environmental interaction outcomes, such as the success or failure of block manipulation actions. This procedurally generated data provides consistent, noise-free annotations, enabling the model to learn precise action-response mappings under diverse and controllable conditions.

\textbf{Curation strategies for high-quality and balanced labeled data.}
Labeled data plays a pivotal role in ensuring model convergence and enhancing controllability.  To ensure that the training data provides strong and reliable supervision signals, we incorporate three key strategies during the construction of labeled Minecraft trajectories using MineRL exploration agents. These strategies are designed to guarantee high visual fidelity, semantic diversity across gameplay scenarios, and balanced distribution across control categories, thereby improving both the robustness and generalization capability of the resulting model.

\textit{Camera motion restriction.}
To ensure viewpoint stability and facilitate the learning of temporally consistent visual representations, we explicitly constrain camera movements during data generation. Specifically, the yaw and pitch angles are restricted to within $15^\circ$ per frame, effectively avoiding sequences with abrupt camera rotations or disorienting viewpoint shifts. Such unstable visual motion, commonly observed in speed, run or chaotic gameplay, can undermine the temporal smoothness of rendered scenes, making it more difficult for the model to learn frame-to-frame coherence and predict stable visual transitions.
By enforcing this constraint at the data creation stage, we produce observation sequences that are visually stable, contextually grounded, and easier to model temporally. This design promotes better alignment between scene perception and control conditioning, ultimately enhancing the model’s ability to generate smoother and more coherent videos in response to user inputs.

\textit{MineRL engine modification.}
To ensure visual consistency and eliminate rendering artifacts, we introduce targeted modifications to the MineRL engine during data creation. Specifically, we disable frustum-based chunk loading, a mechanism that causes new terrain blocks to abruptly appear as the camera moves. This change prevents sudden scene pop-ins that would otherwise disrupt the spatial coherence of the visual stream.
In addition, we implement real-time monitoring of the agent’s health status and in-game interface state. Recording is automatically terminated when the agent is near death, stuck, or when the pause/menu screen is activated. These safeguards ensure that all captured clips reflect uninterrupted, meaningful gameplay interactions rather than irrelevant or low-quality segments.
Together, these modifications help produce high-fidelity video clips that are free of visual artifacts and non-interactive states, providing the model with stable, context-rich supervision signals for learning controllable and temporally coherent world generation.

\begin{table}[t]
    \centering 
    \caption{Biome distributions and characteristics in the balanced \textit{Matrix-Game-MC} dataset for 65-frame controllable video generation. The dataset spans 14 Minecraft biomes, each contributing 4–7\% of samples, ensuring semantic diversity and covering a range of visual and physical conditions.}
    \label{table:balanced_data}
    \begin{threeparttable}
    \scalebox{0.8}{
    \begin{tabular}{l|c|p{8cm}}
        \toprule
        \textbf{Biome} & \textbf{Percentage} & \textbf{Environmental Features} \\
        \midrule
        Forest        & 4.0\%  & Dense trees, wolves, flowers, mushrooms \\
        Taiga         & 4.5\%  & Spruce trees, foxes, berry bushes, snowfall \\
        Swamp         & 4.6\%  & Mangrove trees, slime blocks, lily pads, vines \\
        Ocean         & 7.2\%  & Coral reefs, kelp forests, drowned ruins, prismarine \\
        Mesa          & 6.7\%  & Red sandstone, hardened clay strata, dead bushes \\
        Extreme hills & 7.2\%  & Mountain peaks, emerald ore, snowcaps, waterfalls \\
        Savanna       & 6.3\%  & Baobab trees, acacia wood, herds of llamas \\
        Plains        & 6.0\%  & Rolling grasslands, villages, sunflowers, horses \\
        Beach         & 6.5\%  & Sandy shores, turtle nests, sugarcane, shallow waters \\
        Jungle        & 5.9\%  & Giant trees, cocoa plants, ocelots, temple ruins \\
        River         & 5.8\%  & Flowing water, clay deposits, salmon, gravel banks \\
        Desert        & 7.9\%  & Sand dunes, cacti, desert temples, husks \\
        Mushroom      & 6.3\%  & Mycelium terrain, giant mushrooms, mooshrooms \\
        Icy           & 6.8\%  & Icebergs, polar bears, packed ice, strays \\
        Random        & 14\%   & Random spawn for scenarios like Nether/End  \\
        \bottomrule
    \end{tabular}}
    \end{threeparttable}  
\end{table}

\textit{Scenario diversification.}
We selectively curate 14 Minecraft scenarios, each tied to a specific biome (e.g., forest, desert, ocean), covering diverse terrains, lighting, and structures. Players perform discrete actions (move, jump, attack) in each setting, sampled in a balanced way. This setup exposes the model to varied spatial and physical contexts, improving its ability to generalize motion dynamics and interaction behavior across different virtual world environments.

\textbf{Final labeled dataset for controllable world generation.}
By applying the aforementioned data construction strategies to the MineRL-based exploration agent, we synthesize a high-quality labeled dataset for Minecraft, which forms a substantial portion of our controllable training corpus.  To enhance both visual diversity and control fidelity, we further incorporate procedurally generated videos from Unreal Engine. Together, these sources yield a comprehensive labeled dataset containing over 1,026 hours of video clips for 33-frame training.

To ensure balanced coverage across environments, we additionally curate an expanded dataset comprising more than 1,200 hours of video for 65-frame training. Notably, approximately half of this balanced set is sourced from MineRL-based scenarios spanning 14 distinct Minecraft biomes, such as forest, desert, icy, and mushroom, whose distribution is detailed in Table~\ref{table:balanced_data}.
The resulting datasets are motion-stable, action-dense, and structurally diverse, offering strong and balanced supervision signals for training robust, controllable video generation models capable of generalizing across varied virtual environments.

\section{Matrix-Game: An Interactive World Foundation Model}

\begin{figure}[t]
    \centering
    \includegraphics[width=0.99\textwidth]{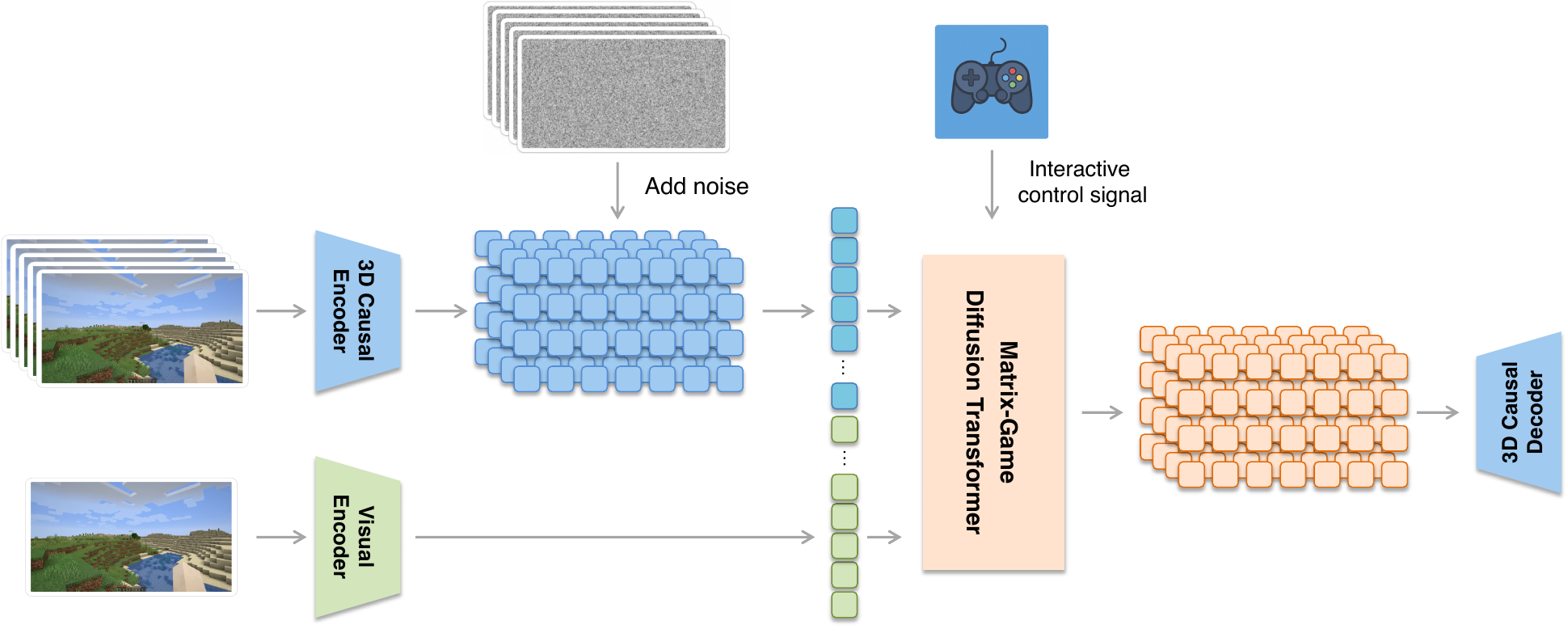}
    \caption{Overview of the interactive image-to-world generation paradigm. The model is trained in a spatiotemporally compressed latent space obtained through a 3D Causal VAE. Conditioned on a reference image along with Gaussian noise and action signals, it generates latent representations that are decoded into video clips.   By grounding generation in the reference image, the model learns to build consistent scene representations that capture geometry, dynamics, and physical interactions, enabling the generation of temporally coherent and spatially structured videos.}
    \label{fig:pipeline}
\end{figure}

\subsection{Model Architecture and Design}  

\paragraph{Exploring spatial intelligence through image-to-world modeling.}
In light of Cosmos~\cite{agarwal2025cosmos} and Genie-2~\cite{parkerholder2024genie2}, we construct our world foundation model, \ie Matrix-Game, using latent diffusion models~\cite{rombach2022high}. Most existing diffusion-based world models, such as SORA~\cite{openai2024worldsim}, HunyuanVideo I2V~\cite{kong2024hunyuanvideo} and Wan~\cite{wang2025wan}, rely on text prompts and reference images as prior knowledge to guide the generation process. While these methods can produce high-quality results, the introduction of text often imposes semantic biases, constrains spatial interpretation, and reduces the model’s ability to ground understanding purely in visual and physical cues. As a result, the model may hallucinate unrealistic content or overfit to language priors rather than faithfully modeling the visual world. Inspired by the concept of Spatial Intelligence~\cite{worldlabs2025generating}, Matrix-Game explores a different route: instead of using both text and images as conditions, our model learns purely from raw images. It learns to understand the world by building a consistent scene that captures geometry, object movement, and how things interact physically. 

As shown in Figure~\ref{fig:pipeline}, Matrix-Game adopts an image-to-world generation paradigm, using a single reference image as the primary prior for world understanding and video generation. The model is trained in a spatiotemporally compressed latent space constructed by a 3D Causal VAE~\cite{kingma2013auto,yu2023language}, which reduces the spatial and temporal resolution of video sequences by factors of 8 and 4, respectively. The reference image, processed by either a visual encoder or a multi-modal backbone, serves as the central conditioning input. Conditioned on Gaussian noise and optional user actions, a Diffusion Transformer (DiT) generates latent representations, which are then decoded into coherent video sequences via the 3D VAE decoder.

Unlike prior methods that treat the image as just an auxiliary input alongside text prompts, Matrix-Game is trained to perceive, interpret, and model the world directly from visual content alone. Through large-scale training on diverse game-world environments, the model learns to recognize the spatial layout of scenes, how objects move, and how they physically interact. This allows it to build a consistent and structured understanding of the world from a single image, which forms the basis for generating realistic and controllable videos.

By combining visual understanding with user control, Matrix-Game transforms video generation into an interactive tool for exploration and creation, empowering users to see, modify, and build coherent virtual worlds starting from a single image.

\begin{figure}[t]
    \centering
    \includegraphics[width=1\textwidth]{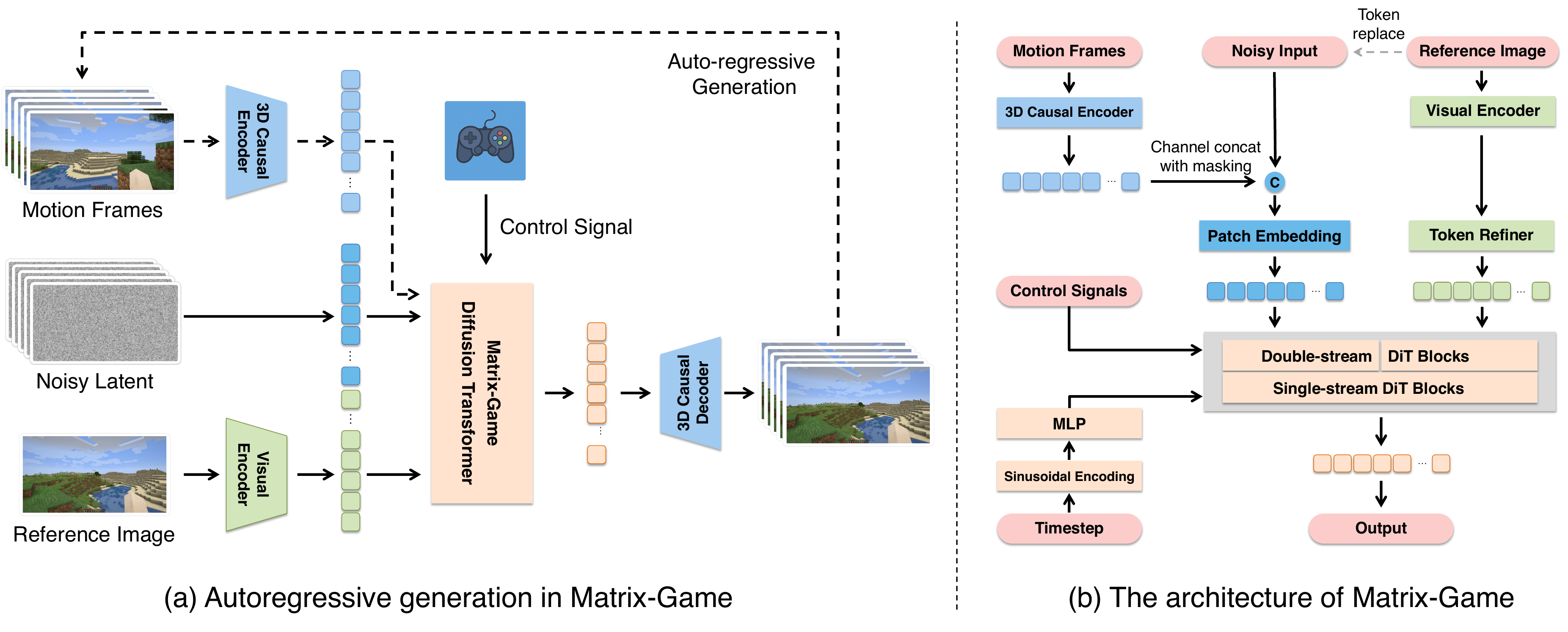}
    \caption{(a) Autoregressive generation in Matrix-Game and (b) The architecture of  Matrix-Game. To enable long-duration video generation, Matrix-Game adopts an autoregressive strategy: the last few frames of each generated clip are used as motion conditions for generating the next clip. Specifically, the latent of these motion frames are concatenated with the noisy latent along the channel dimension, and a binary mask is also concatenated to indicate which frames contain valid motion information. This design enhances local temporal consistency across video segments, allowing the model to maintain coherent dynamics over extended time horizons. Moreover, we adopt the token replacement trick  in HunyuanVideo I2V~\cite{kong2024hunyuanvideo} to enable stable image-to-video generation.}
    \label{fig:autoregressive_architecture}
\end{figure}

\paragraph{Autoregressive generation with diffusion transformer.}  To enable high-quality generation, following Flux~\cite{flux2024} and  HunyuanVideo~\cite{kong2024hunyuanvideo}, we adopt a multi-modal diffusion transformer (MMDiT) for image-to-world modeling. 
Like most current video generation methods~\cite{openai2024worldsim,kong2024hunyuanvideo,wang2025wan}, the image-to-world model, as illustrated in Figure~\ref{fig:pipeline}, generates fixed-length video clips,  which limits its applicability in real-world scenarios that demand long-term or continuous world modeling. To address this limitation, and drawing inspiration from recent advances in long-duration video generation~\cite{tian2024emo,zheng2024memo}, we adopt an autoregressive strategy: at each step, the model takes the previously generated video clip as motion context to produce the next. As shown in Figure~\ref{fig:autoregressive_architecture}(a), we use the last $k=5$ frames from each generated segment as motion conditions for generating the subsequent clip. Such a design enables the model to progressively extend generation over time while preserving temporal coherence across segments.

To this end, as illustrated in Figure~\ref{fig:autoregressive_architecture}(b), we concatenate the latent representations of the motion frames with the noisy latent along the channel dimension to form the input for the next generation step. A binary mask is concatenated to indicate which frames contain valid motion information. The combined latent tensor is then processed through a patch embedding layer and further concatenated with the image tokens along the token dimension. Finally, conditioned on user control signals as an additional guidance input, the multi-modal diffusion transformer generates a new video clip.

However, a key challenge in autoregressive generation is temporal error accumulation: artifacts in the last few generated frames can propagate and amplify in subsequent segments. To improve the robustness of the autoregressive process, inspired by the Open-Sora plan~\cite{lin2024open}, we introduce Gaussian noise to the motion frames and reference images with a probability of 0.2 during training. In addition, we apply classifier-free guidance (CFG) to  motion frames during training: the latents of motion frames are replaced with unconditioned signals (\ie zero latents) with a probability of 0.25. This CFG strategy encourages the model to rely more effectively on motion context, leading to more stable and reliable autoregressive video generation.

\begin{figure}[t]
    \centering
    \includegraphics[width=0.99\textwidth]{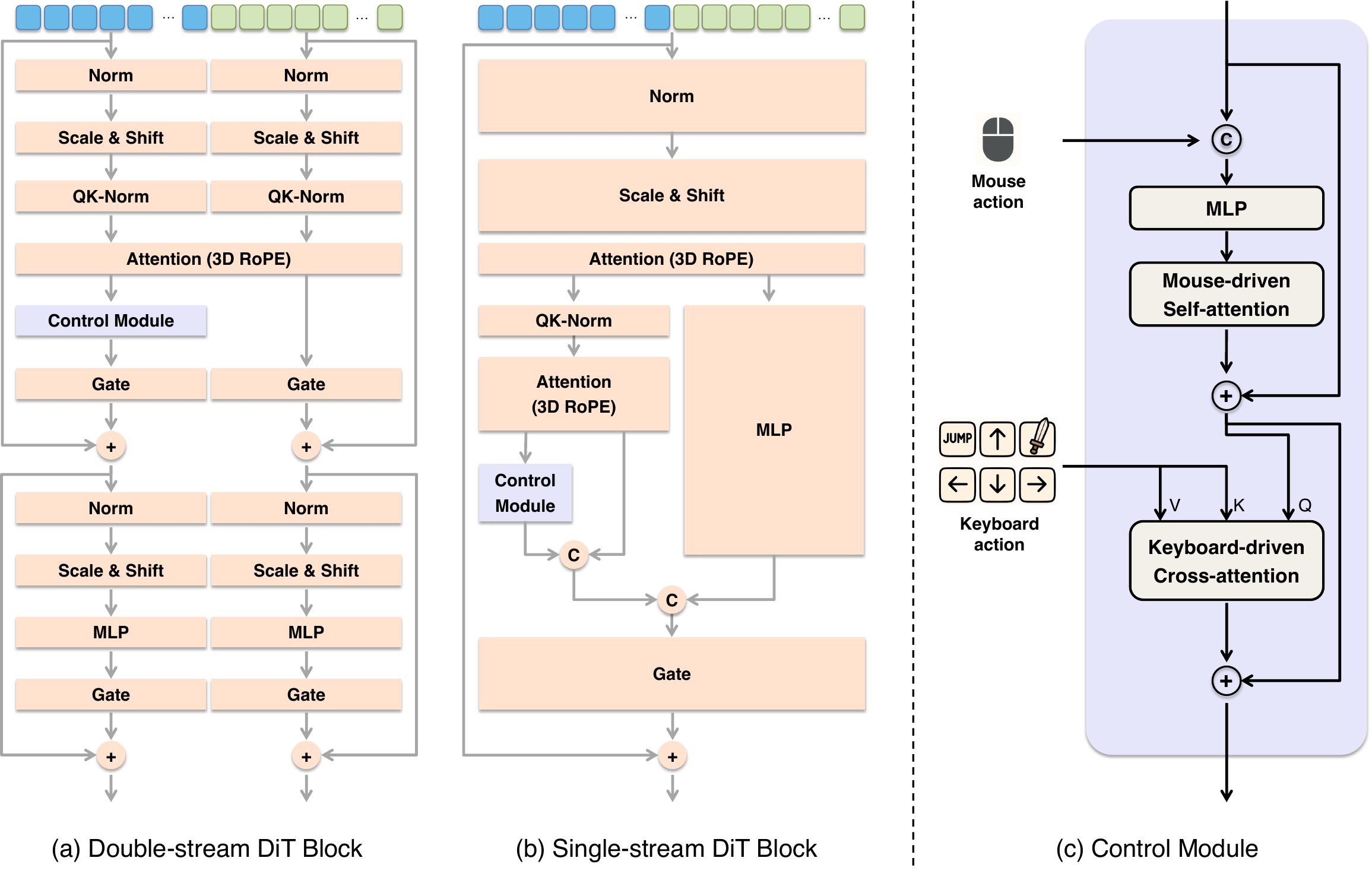}
    \caption{The details of diffusion transformer blocks in Matrix-Game.}
    \label{fig:block_architecture}
\end{figure}

\paragraph{Injecting actions for controllable video generation.}
Inspired by Genie 2~\cite{parkerholder2024genie2}, we employ frame-level control signals to guide video generation. As shown in Figures~\ref{fig:block_architecture}(a-b), we integrate a control module into our multi-modal diffusion transformer to enable action-controllable generation, following the action control design introduced in GameFactory~\cite{yu2025gamefactory}. The detailed architecture of this action control module is shown in Figure~\ref{fig:block_architecture}(c). 

Specifically, we use discrete encodings to represent keyboard actions, including “up,” “down,” “left,” “right,” “jump,” and “attack”, and continuous scalar values to represent mouse movements, defined as changes in pitch angle. To align these action signals with the compressed latent tokens produced by the 3D Causal VAE, we adopt the group operation trick~\cite{yu2025gamefactory}, which accounts for the temporal compression ratio and enables effective token-level conditioning. The continuous mouse action is concatenated with the input latent and processed through an MLP followed by temporal self-attention, while the discrete keyboard actions are integrated via cross-attention to guide the diffusion process.

We also apply classifier-free guidance to action signals during training by replacing them with unconditioned signals with a probability of 0.1. This helps the model learn to use action signals more effectively when they are provided, resulting in better control and interaction in the generated videos.

\subsection{Model Training}
To improve training stability and enable faster inference, Matrix-Game leverages the flow matching paradigm~\cite{esser2024scaling}, which outperforms traditional denoising diffusion probabilistic models (DDPM)\cite{ho2020denoising} in both convergence and sampling efficiency. Based on this paradigm, we adopt the rectified flow loss~\cite{liu2023flow} for model training. To support complex tasks such as modeling world knowledge, capturing physical dynamics, and enabling action-controllable generation, we organize the training process into two progressive stages, each optimized for distinct learning objectives.

\paragraph{Stage 1: Unlabeled training for game world understanding.} 
To accelerate convergence, the model is initialized with pretrained weights from the HunyuanVideo image-to-video model~\cite{kong2024hunyuanvideo}.
To shift from text-driven generation to image-conditioned world modeling, we replace the original text branch in the multi-modal diffusion transformer with an image branch, as illustrated in Figure~\ref{fig:autoregressive_architecture}(b). The action control module is excluded in this stage to focus on visual world understanding.

The primary goal is to pretrain the model on large-scale game environments, enabling it to build structured understanding of virtual worlds, including spatial layouts, object dynamics, and intuitive physical rules. To this end, we use 2,700-hour unlabeled Minecraft videos at 720p resolution as a rich source of visual and physical cues. We train the model using a diverse mixture of frame counts (17, 33, and 65) and aspect ratios (16:9, 4:3, and 21:9) to enhance robustness under varied temporal and spatial settings.

Following the initial large-scale pretraining, we further refine the model’s visual and physical understanding by curating a subset of 870-hour high-quality video clips from the same dataset. These clips are selected based on stable camera motion, clean user interfaces, and overall visual clarity. This targeted refinement improves Matrix-Game’s ability to model coherent spatial structures, capture fine-grained physical interactions, and generate videos with higher perceptual quality and temporal consistency.

\paragraph{Stage 2: Action-labeled training for interactive world generation.}
In the second stage, we integrate the action control module into a multi-modal diffusion transformer to enable action-controllable video generation. The final model, Matrix-Game, contains 17 billion parameters and is trained on 1,200 hours of action-labeled 720p 33-frame video clips collected from both Minecraft and Unreal Engine environments. To ensure training stability and efficiency during controllable generation, we adopt a fixed 720p resolution and a 33-frame setting in the early training phase.

To mitigate class imbalance in world scenarios~\cite{zhang2023deep,zhang2022self}, we further refine the training data in the second sub-stage. Specifically, we curate a more balanced dataset by organizing samples across 8 distinct Minecraft biomes: beach, desert, forest, hills, icy, mushroom, plains, and river. Combined with procedurally generated data from Unreal Engine, this results in a high-quality, balanced training set comprising approximately 1,200 hours of 720p 65-frame video clips.

We continue training under the 65-frame setting to strengthen the model’s ability to capture long-range temporal dependencies, which are essential for maintaining coherent interaction across extended sequences. Through this integration of balanced, action-rich data and strong visual priors, Matrix-Game learns to precisely interpret user inputs and generalize across diverse interactive environments. This tight coupling of visual understanding and user control advances video generation into an interactive paradigm for world exploration and creation, empowering users to perceive, modify, and construct coherent virtual environments from a single reference image.

\section{GameWorld Score: A Unified Benchmark for Minecraft World Models}
\label{sec:gameworld_benchmark}

With the rise of world models, an increasing number of studies have focused on the Minecraft world generation, aiming to leverage video generation models to produce videos that not only align with user action inputs, but also adhere to the physical rules inherent in the game. However, existing research lacks a unified evaluation benchmark to consistently measure and compare model performance in the setting with action inputs. Current benchmarks, such as VBench~\cite{huang2024vbench}, primarily evaluate visual quality and the alignment between generated videos and textual prompts. {WorldScore}~\cite{duan2025worldscore} goes a step further by assessing the ability of models to generate 3D/4D worlds conditioned on textual inputs. However, these benchmarks remain limited in evaluating image-to-world generation, and are inadequate for scenarios involving fine-grained control signals, such as frame-by-frame action conditions commonly found in real gameplay settings.

Minecraft world generation involves a combination of {discrete control signals} (e.g., move forward, move backward, move left, move right, jump, attack, \etc) and {continuous control inputs} (e.g., camera rotation along the X and Y axes). Therefore, a comprehensive benchmark needs to accommodate these types of controls and offer corresponding evaluations. Furthermore, Minecraft follows fundamental physical laws, such as  \textit{object consistency} and \textit{scenario consistency}. A capable world model should generate outputs that are consistent with these physical intuitions.

To better measure and compare Minecraft world models, we develop \textbf{GameWorld Score}, a unified benchmark that evaluates not only the perceptual quality of generated videos, but also their \textit{controllability} and \textit{physical plausibility}. Specifically, we decompose the evaluation of world model performance into {eight dimensions}, each targeting a distinct aspect of video generation. At the top level, GameWorld evaluates models across four key pillars:

\textbf{Visual quality:} Assesses each individual frame's visual fidelity in alignment with the Human Visual System (HVS), focusing on clarity, coherence, and realism of still images.

\textbf{Temporal quality:} Measures how well the model maintains consistency and smoothness over time, capturing dynamics like motion continuity and temporal coherence.

\textbf{Action controllability:} Evaluates whether the generated video faithfully follows the user-provided control inputs, such as movement commands and camera adjustments.
 
\textbf{Physical rule understanding:} Evaluates whether the video adheres to fundamental physical principles, with a particular focus on maintaining object consistency and scenario consistency across space and time, reflecting the model's ability to simulate physically coherent environments.

\begin{figure}[t]
    \centering
    \vspace{-0.2in}
    \includegraphics[width=0.8\linewidth]{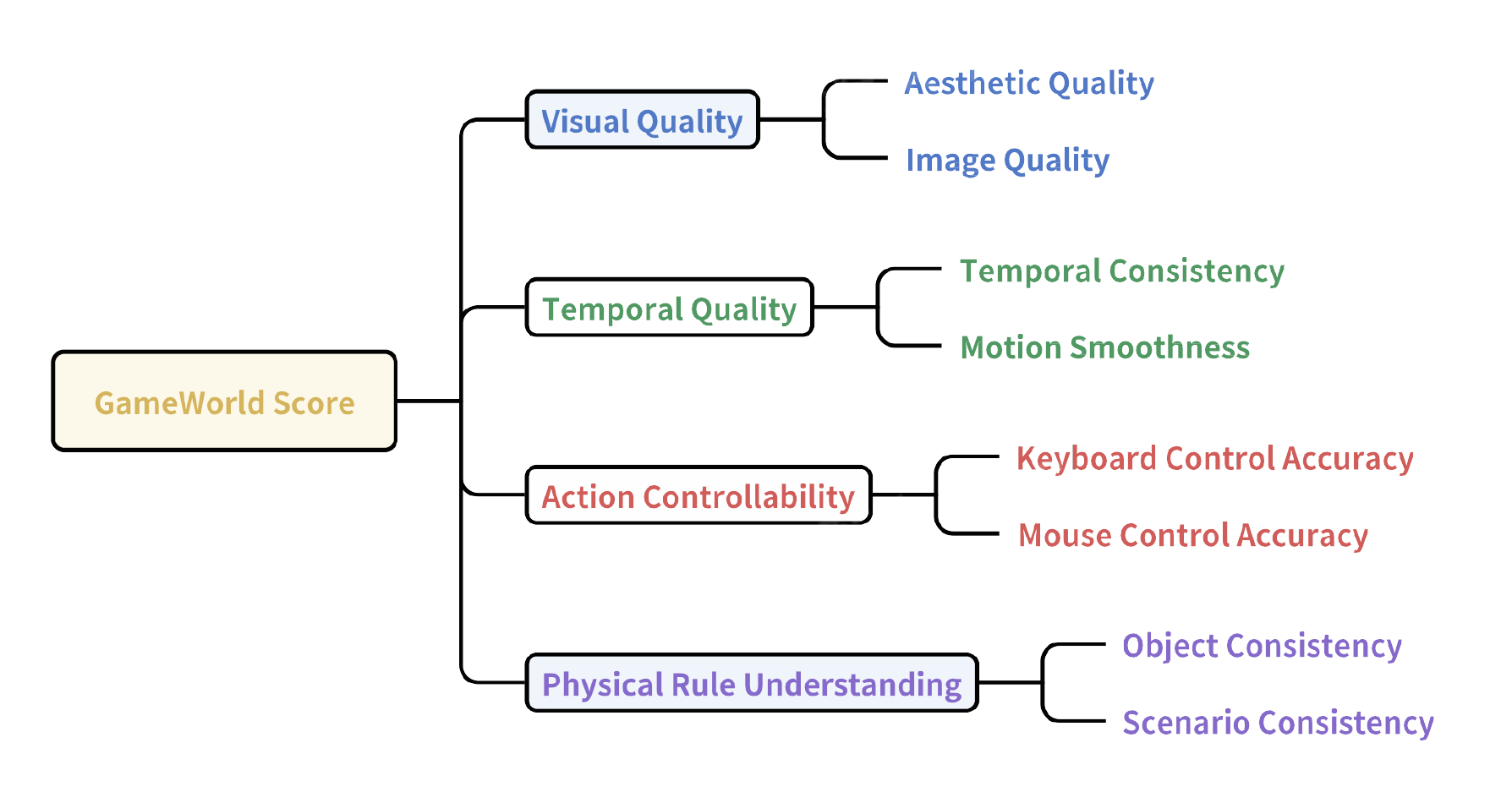}
    \vspace{-0.1in}
    \caption{GameWorld Score provides a unified benchmark for assessing the quality and realism of generated Minecraft worlds.}
    \vspace{-0.1in}
    \label{fig:benchmark}
\end{figure}

These four pillars are further divided into fine-grained dimensions, as illustrated in Figure~\ref{fig:benchmark}, enabling a comprehensive and structured assessment of generative models within interactive and physics-driven environments like Minecraft.

\subsection{Visual Quality}
\label{subsec:frame_quality}

Frame-wise quality measures the merit of \emph{individual} frames, ignoring their interplay across time.  
We analyse every frame from two complementary perspectives:  

\noindent\textbf{Aesthetic quality.}  
We evaluate the perceived visual appeal of individual frames using the LAION aesthetic predictor~\cite{LAIONaes}, a learned model trained on large-scale human aesthetic preferences. This score reflects a combination of factors, including image composition, color harmony, lighting balance, photorealism, and stylistic coherence. A higher aesthetic score indicates stronger alignment with human judgments of visual attractiveness, providing a complementary perspective beyond pixel-level or structural metrics.

\noindent\textbf{Image quality.}  To assess the perceptual fidelity of each frame, we evaluate low-level visual artifacts such as over-exposure, noise, compression distortions, and blurriness using the MUSIQ predictor~\cite{Ke2021MUSIQ}. MUSIQ is a no-reference image quality assessment model trained on the SPAQ dataset~\cite{Fang2020spaq}, which reflects diverse real-world imaging conditions. This metric provides a quantitative measure of how clean, sharp, and artifact-free the generated frames appear, serving as a robust proxy for human-perceived visual quality.

\subsection{Temporal Quality}
\label{subsec:temporal_quality}

Temporal Quality evaluates how well the generated video maintains consistency and realism across consecutive frames. This aspect is crucial for ensuring coherent motion, avoiding flickering artifacts, and preserving object integrity over time. To capture different facets of temporal stability, we propose two complementary dimensions:

\noindent\textbf{{Temporal consistency.}  }
To evaluate how well the background and scene remain stable over time, we compute the pairwise similarity of CLIP~\cite{radford2021clip} features extracted from each frame in the video sequence. CLIP embeddings capture high-level semantic and visual information, making them suitable for assessing whether consecutive frames depict a temporally coherent scene. Specifically, we calculate the average cosine similarity between adjacent frames to quantify consistency. Higher similarity indicates that the model preserves static elements such as background layout across time, thus avoiding common artifacts like flickering, texture drift, or abrupt visual changes.

\noindent\textbf{{Motion smoothness.}  }
While temporal consistency ensures stable appearance across frames, it does not account for the quality of motion itself. Abrupt or jittery transitions may still occur even when frame content appears coherent. To address this, we assess motion smoothness by evaluating whether the movement of objects and camera follows physically plausible and temporally continuous trajectories. Following~\cite{licvpr23amt}, we leverage the motion priors learned by a video frame interpolation network to detect unnatural dynamics. Specifically, we feed the generated video into a pretrained interpolation model~\cite{licvpr23amt} and measure the reconstruction error between actual frames and those interpolated from adjacent frames. High interpolation accuracy implies that motion flows smoothly between frames, while large discrepancies indicate irregularities such as jitter, stuttering, or frame-level discontinuities. This method provides a proxy for judging motion realism without requiring dense annotations.  

\subsection{Action Controllability}

This module aims to assess how well the generated videos understand and follow the input action conditions. Ideally, the visual content of generated videos should respond frame-by-frame to the given player-like control signals, mirroring the behavior of an interactive game environment. In our setup, we categorize the input signals into two types: \textbf{keyboard conditions} and \textbf{mouse conditions}. Keyboard conditions govern directional movement (forward, back, left, right) as well as jump and attack actions. Mouse conditions control the camera orientation in eight directions (up, down, left, right, upper-left, upper-right, lower-left, and lower-right), represented by rotational angle changes.

Following MineWorld~\cite{guo2025mineworld}, we adopt the \textit{Inverse Dynamics Model (IDM)} to assess controllability by inferring the underlying action conditions from a given video sequence. Controllability is measured by comparing the inferred actions with ground truth inputs, evaluating how accurately the generated video reflects intended control signals. Please note that IDM, trained on 1,962 hours of Minecraft gameplay, achieves 90.6\% accuracy on keyboard prediction and an $R^2$ score of 0.97 for mouse movement regression, making it a reliable proxy for extracting action labels from video.

\noindent\textbf{Keyboard control accuracy.} 
We evaluate the controllability of keyboard inputs by computing precision across four grouped action categories: (forward, back, empty), (left, right, empty), (attack, empty), and (jump, empty). Each group is treated as a multi-class classification problem with mutually exclusive actions. The final keyboard condition accuracy is reported as the average precision across these four groups. In addition to this aggregated score, we also report the per-class precision for each individual action (e.g., forward, left, jump, etc.), which provides a more fine-grained analysis of how well the model responds to different types of control inputs. This allows us to analyze how well the model follows specific control commands.

\noindent\textbf{Mouse control accuracy.} 
Mouse inputs affect the camera's rotational movement, which is modeled independently from keyboard actions. For each axis (x and y), directional movement is detected when the absolute rotational change exceeds a pre-defined threshold. This results in nine categories: up, down, left, right, upper-left, upper-right, lower-left, lower-right, and empty. A prediction is considered correct if the motion direction in the generated video matches the labeled condition. The final accuracy is reported as the precision over all positive predictions.

\subsection{Physical Rule Understanding}
To assess the model’s understanding of the physical world rule, we evaluate its ability to preserve {Object Consistency} and Scenario Consistency across frames.

\noindent\textbf{{Object Consistency.}} A physically grounded model should preserve object geometry over time despite texture changes. Following WorldScore~\cite{duan2025worldscore}, we use DROID-SLAM~\cite{teed2021droid} to estimate depth and camera pose, and compute the \textit{reprojection error} between co-visible pixels across frames. Since DROID-SLAM is robust to appearance changes, this metric isolates geometric consistency at the object level. Lower error indicates better preservation of object structure across time.

\textbf{Scenario Consistency.}
We propose a Scenario Consistency metric to evaluate how well a model preserves the overall scene over time. It uses 8 types of symmetric camera motion—e.g., up-then-down, left-then-right, and diagonal pairs—where the camera moves in one direction and then reverses along the same path. Corresponding frames in each direction should ideally match. The motion is large enough to move the scene out of view, forcing the model to recover it during the return. Consistency is measured by the mean-squared error (MSE) between paired frames, allowing up to 4-pixel shifts to handle minor misalignments. Lower error indicates better scenario consistency.

\begin{table}[t]
\vspace{-0.2in}
\centering
\caption{Comparisons of Minecraft world generation in terms of the GameWorld Score benchmark. {Keyboard Acc.} and {Mouse Acc.} represent the accuracy of control signal prediction for keyboard actions and mouse-driven camera movements, respectively. {Motion smooth.} evaluates the smoothness of frame-wise motion transitions, while {Temporal Cons.} measures short-range temporal consistency. {Obj. Cons.} reflects the 3D
consistency of objects across space and time, while Scenario Cons. indicates the the consistency of scenarios across time.}
\label{table:exp_result}
\begin{threeparttable}
\scalebox{0.555}{
\begin{tabular}{l|*{12}{c}}
\toprule
\multirow{2}{*}{Model} & \multicolumn{2}{c}{Visual Quality} && \multicolumn{2}{c}{Temporal Quality} && \multicolumn{2}{c}{Action Controllability} && \multicolumn{2}{c}{Physical Understanding} \\
\cmidrule{2-3}\cmidrule{5-6}\cmidrule{8-9}\cmidrule{11-12}
 & Image Quality $\uparrow$ & Aesthetic $\uparrow$ && Temporal Cons. $\uparrow$ & Motion smooth. $\uparrow$ && Keyboard Acc. $\uparrow$ & Mouse Acc. $\uparrow$ && Obj. Cons. $\uparrow$ & Scenario Cons. $\uparrow$ \\
\midrule
Oasis~\cite{oasis2024} & 0.65 & 0.48 && 0.94 & \textbf{0.98} && 0.77 & 0.56 && 0.56 & 0.86 \\
MineWorld~\cite{guo2025mineworld} & 0.69 & 0.47 && 0.95 & \textbf{0.98} && 0.86 & 0.64 && 0.51 & 0.92 \\
Ours & \textbf{0.72} & \textbf{0.49} && \textbf{0.97} & \textbf{0.98} && \textbf{0.95} & \textbf{0.95} && \textbf{0.76} &\textbf{0.93} \\
\bottomrule
\end{tabular}
\vspace{-0.2in}
}
\end{threeparttable}
\end{table}

\section{Experiments}

\textbf{Experimental objectives}.  Our experiments are designed to comprehensively evaluate the proposed model across multiple dimensions. Specifically, we aim to answer the following four questions:
(1) {GameWorld Score Benchmark:} Does our model outperform existing state-of-the-art open-source Minecraft models across key dimensions such as visual quality, temporal quality, action controllability, and  physical rule understanding?
(2) {Action Controllability:} How well does our model respond to various user commands, particularly keyboard actions and mouse movements?
(3) {Scene Generalization:} How does the model perform across diverse Minecraft scenarios (e.g., forest, desert, icy, mushroom)?
(4) {Autoregressive Generation:} Can our model maintain coherent and controllable behavior during long-horizon autoregressive video generation?

\textbf{Implementation details}. 
The experiments were conducted with a per-GPU batch size of 1. We employed bf16 mixed precision and the Fully Sharded Data Parallel (FSDP) strategy for efficient large-scale training. The learning rate was set to $5 \times 10^{-5}$, with 16 training FPS and 5 motion frames. During inference, we applied Classifier-Free Guidance (CFG) to the reference image, motion frames, and action signals. The CFG scale was set to 6, with 50 sampling steps using Flow Matching. The flow matching shift parameter was set to 15.

\textbf{Compared methods}. 
To establish a solid comparison, we include two of the most representative open-source world models as baselines: OASIS~\cite{oasis2024} and MineWorld~\cite{guo2025mineworld}. Both are recent works released with code and models available, and have shown competitive results in Minecraft world generation. These models provide a reasonable benchmark for evaluating visual quality, temporal dynamics, and controllability, allowing us to position our method in relation to existing publicly available systems.

\textbf{Evaluation metrics}. We evaluate Minecraft world generation performance using the proposed {GameWorld Score} benchmark (cf. Section~\ref{sec:gameworld_benchmark}). Moreover, to complement standard quantitative metric, which often fail to capture subtle differences in perceptual quality, we further conduct human evaluation through manual scoring on all baseline outputs. This evaluation covers four key aspects: \emph{Overall Quality}, \emph{Controllability}, \emph{Visual Quality}, and \emph{Temporal Consistency}. It is performed in a {double-blind setting} with {two independent annotator groups}, where neither group is aware of the method identities, ensuring fairness and minimizing potential bias.

\subsection{Model Performance}
\textbf{GameWorld Score benchmark.} As shown in Table~\ref{table:exp_result} and Figure~\ref{fig:radar_1}, {Matrix-Game} achieves strong and well-rounded performance across all dimensions of the {GameWorld Score} benchmark. Compared to leading open-source baselines such as OASIS~\cite{oasis2024} and MineWorld~\cite{guo2025mineworld}, our model excels especially in two key aspects: {controllability} and {physical consistency}. It responds more accurately to control inputs, including both keyboard and mouse signals, which is critical for interactive world generation. Additionally, it better preserves the object and scenarios of the game world over time, demonstrating a stronger understanding of physical layout and geometry. 
Beyond these, our {Matrix-Game} also maintains competitive performance in other key aspects. It produces sharper and more aesthetically pleasing visuals, as reflected in both the {image quality} and {aesthetic} scores. Furthermore, it achieves high {temporal consistency} and {motion smoothness}, ensuring seamless transitions between frames without flickering or abrupt changes. Altogether, these results highlight the model’s ability to deliver high-quality, physically grounded, and user-controllable video outputs suitable for complex interactive world generation.

\begin{figure}[t]
    \centering     \vspace{-0.2in}
    \includegraphics[width=0.7\linewidth]{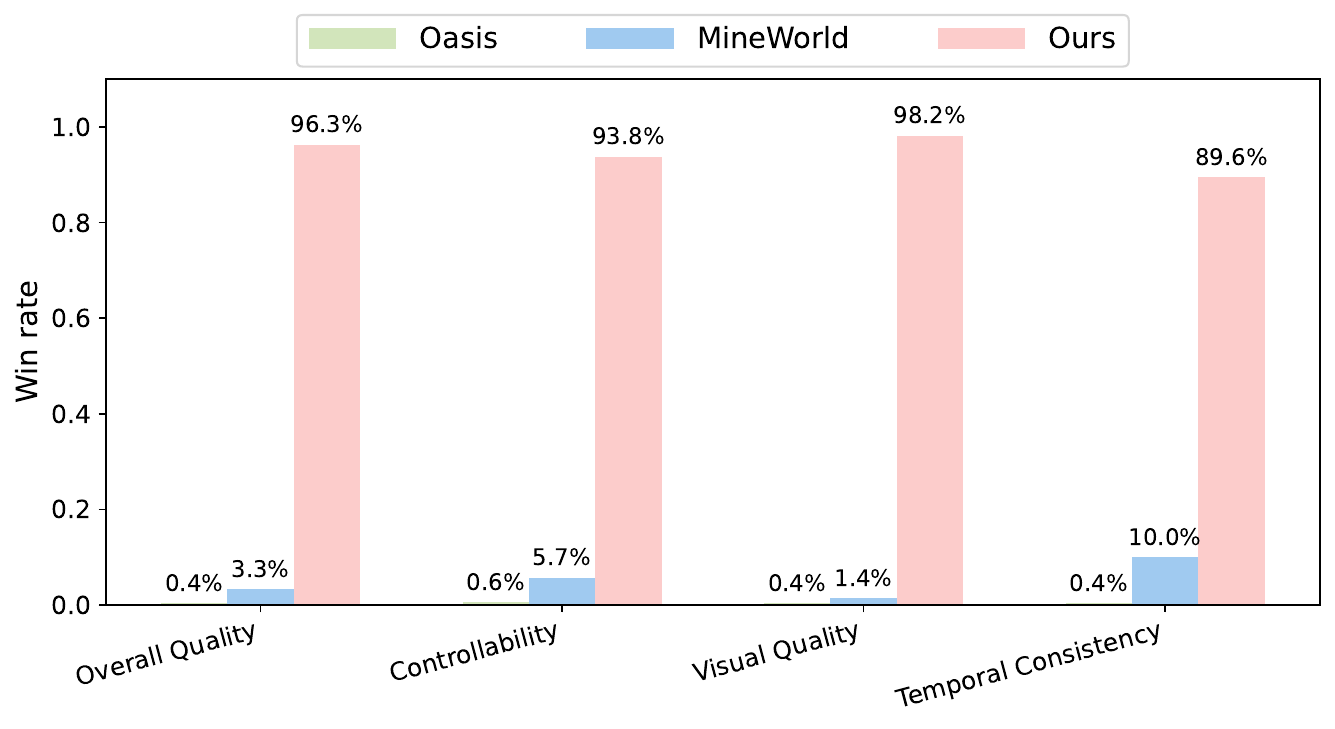}
    \vspace{-0.1in}
    \caption{Human evaluation results (double-blind setting) across four predefined dimensions: Overall Quality, Controllability, Visual Quality, and Temporal Consistency. The win rate reflects the proportion of scenario-metric pairs in which each method is rated as the best by annotators.}  
    \vspace{-0.1in}
    \label{fig:human_evaluation}
\end{figure}

\textbf{Human studies.} To complement objective metrics and mitigate potential biases, we conducted human evaluation through two independent double-blind studies. Each study was carried out by a separate group of annotators, who were unaware of the method identities and each other’s assessments. Annotators compared video outputs across four key dimensions: Overall Quality, Controllability, Visual Quality, and Temporal Consistency. As shown in Figure~\ref{fig:human_evaluation}, our method is overwhelmingly favored by human evaluators across all assessed dimensions. It achieves a 96.30\% win rate in Overall Quality, indicating that users consistently preferred our results when evaluating the overall impression of the generated videos, including realism, coherence, and completeness. In terms of Controllability, our method scores 93.76\%, reflecting its strong ability to accurately translate keyboard and mouse inputs into desired in-game behaviors. This highlights its superiority in generating interactive content with precise user control. Our approach also achieves 98.23\% in Visual Quality, showcasing clear advantages in perceptual fidelity, texture clarity, and aesthetic appeal of individual frames. Finally, with 89.56\% in Temporal Consistency, our method demonstrates robust temporal coherence, ensuring that motion remains stable and physically plausible across frames. Importantly, these human evaluation results, based solely on subjective preference, align with our proposed {GameWorld Score} benchmark, reinforcing its reliability as an effective proxy for human judgment.

\begin{table}[t]
\centering
\caption{Comparison of control accuracy. The evaluation covers two types of control signals: {keyboard actions} and {camera movements}. Camera movement is categorized into eight directions based on pitch and yaw angle changes: $\uparrow$, $\downarrow$, $\leftarrow$, $\rightarrow$, and their diagonals ($\nwarrow$, $\nearrow$, $\swarrow$, $\searrow$). Accuracy is computed using an Inverse Dynamics Model (IDM) by comparing predicted actions from generated videos with ground truth labels. Higher accuracy values represent better controllability.}
\label{tab:control_signals}
\begin{threeparttable}
\scalebox{0.55}{
\begin{tabular}{l|*{15}{c}}
\toprule
\multirow{2}{*}{Model} & \multicolumn{6}{c}{Keyboard Action} && \multicolumn{8}{c}{Mouse Movement Action} \cr\cmidrule{2-7}\cmidrule{9-16}   
 & forward & backward & left & right & jump & attack && camera$\uparrow$ & camera$\downarrow$ & camera$\leftarrow$ & camera$\rightarrow$ & camera$\nwarrow$ & camera$\nearrow$ & camera$\swarrow$ & camera$\searrow$ \\
\midrule
Oasis~\cite{oasis2024} & 0.85 & 0.78 & 0.80 & 0.79 & 0.77 & 0.89 && 0.66 & 0.55 & 0.33 & 0.35 & 0.56 & 0.53 & 0.45 & 0.51 \\

MineWorld~\cite{guo2025mineworld} & 0.86 & 0.80 & 0.87 & 0.88 & 0.82 & 0.87 && 0.46 & 0.45 & 0.53 & 0.54 & 0.66 & 0.77 & 0.87 & 0.96 \\

Ours & \textbf{0.99} & \textbf{0.91} & \textbf{0.92} & \textbf{0.96} & \textbf{0.88} & \textbf{0.95} && \textbf{0.91} & \textbf{0.98} & \textbf{0.89} & \textbf{0.90} & \textbf{0.92} & \textbf{0.97} & \textbf{0.98} & \textbf{0.98} \\
\bottomrule
\end{tabular}
}
\end{threeparttable}
\end{table}

\subsection{Action Controllability} 
\begin{figure}[p]
\centering
\begin{subfigure}{0.99\linewidth}
    \centering
    \includegraphics[width=\linewidth]{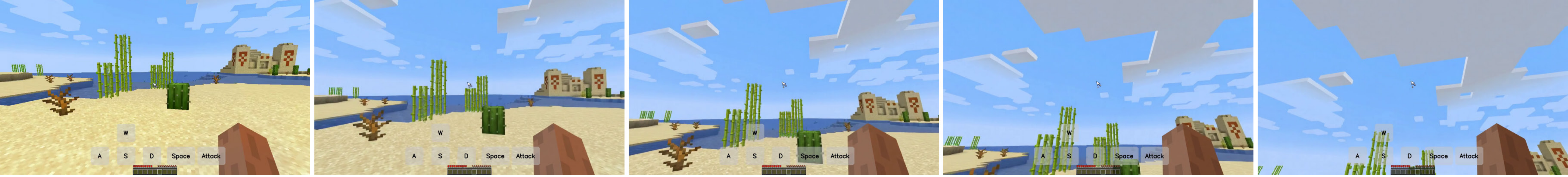}
    \caption{Camera $\uparrow$}
    \vspace{0.1in}
\end{subfigure}
\begin{subfigure}{0.99\linewidth}
    \centering
    \includegraphics[width=\linewidth]{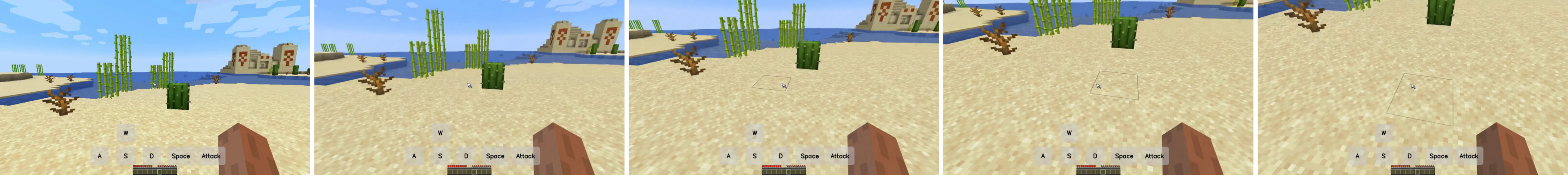}
    \caption{Camera $\downarrow$}
    \vspace{0.1in}
\end{subfigure}  

\begin{subfigure}{0.99\linewidth}
    \centering
    \includegraphics[width=\linewidth]{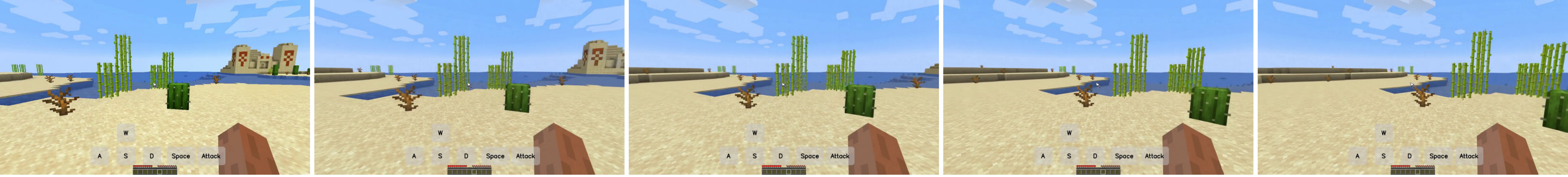}
    \caption{Camera $\leftarrow$}
    \vspace{0.1in}
\end{subfigure}  

\begin{subfigure}{0.99\linewidth}
    \centering
    \includegraphics[width=\linewidth]{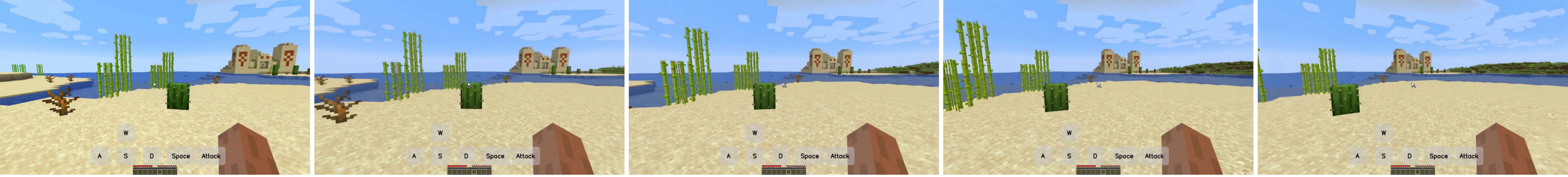}
    \caption{Camera $\rightarrow$}
    \vspace{0.1in}
\end{subfigure}  

\begin{subfigure}{0.99\linewidth}
    \centering
    \includegraphics[width=\linewidth]{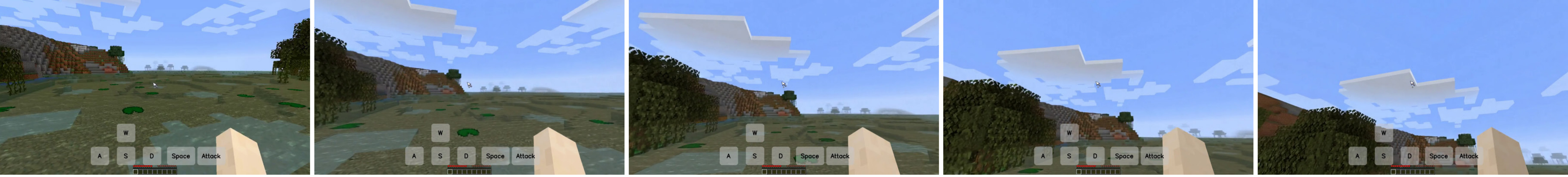}
    \caption{Camera $\nwarrow$}
    \vspace{0.1in}
\end{subfigure}  

\begin{subfigure}{0.99\linewidth}
    \centering
    \includegraphics[width=\linewidth]{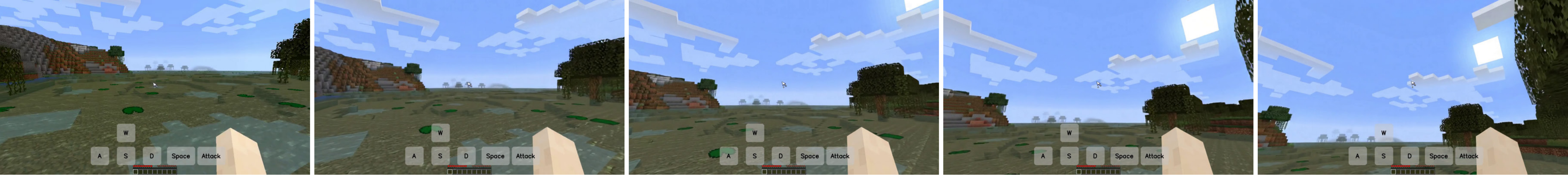}
    \caption{Camera $\nearrow$}
    \vspace{0.1in}
\end{subfigure}  

\begin{subfigure}{0.99\linewidth}
    \centering
    \includegraphics[width=\linewidth]{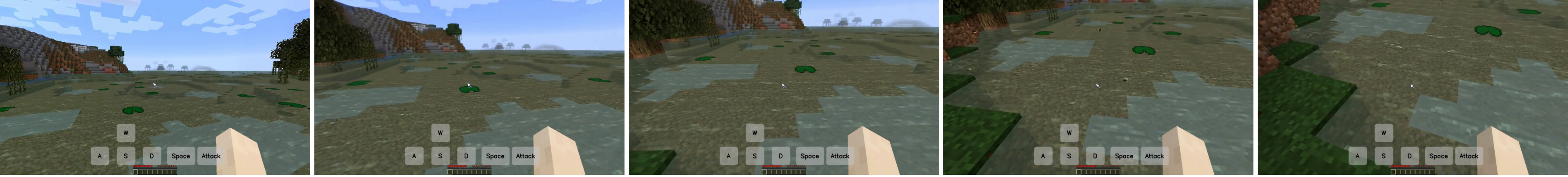}
    \caption{Camera $\swarrow$}
    \vspace{0.1in}
\end{subfigure}  

\begin{subfigure}{0.99\linewidth}
    \centering
    \includegraphics[width=\linewidth]{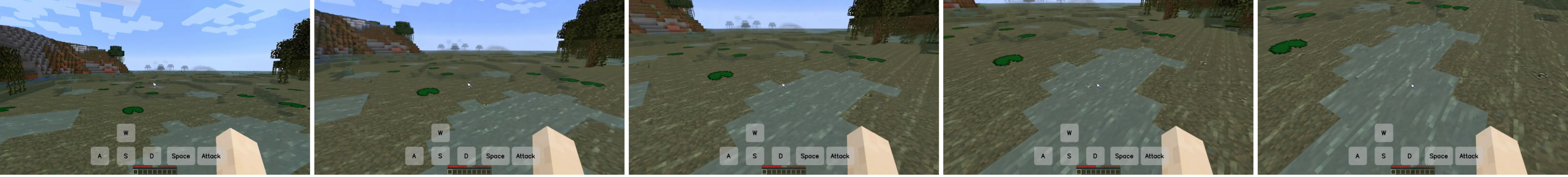}
    \caption{Camera $\searrow$} 
\end{subfigure}  
\caption{Demonstration of \textit{Matrix-Game}'s controllable video generation conditioned on diverse mouse movement commands, including horizontal (left, right), vertical (up, down), and diagonal camera adjustments. The model responds accurately to subtle changes in camera direction and produces geometrically consistent world views aligned with user control signals.}
\label{fig:demo_mouse_actions}
\end{figure}

\begin{figure}[p]
\centering
\begin{subfigure}{0.99\linewidth}
    \centering
    \includegraphics[width=\linewidth]{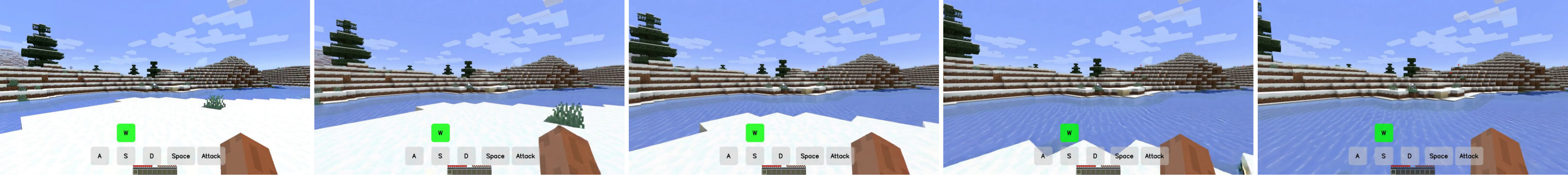}
    \caption{Forward}
    \vspace{0.1in}
\end{subfigure}
\begin{subfigure}{0.99\linewidth}
    \centering
    \includegraphics[width=\linewidth]{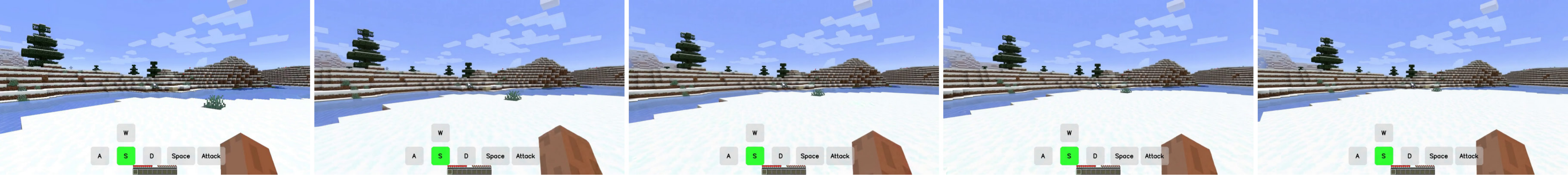}
    \caption{Back}
    \vspace{0.1in}
\end{subfigure}  

\begin{subfigure}{0.99\linewidth}
    \centering
    \includegraphics[width=\linewidth]{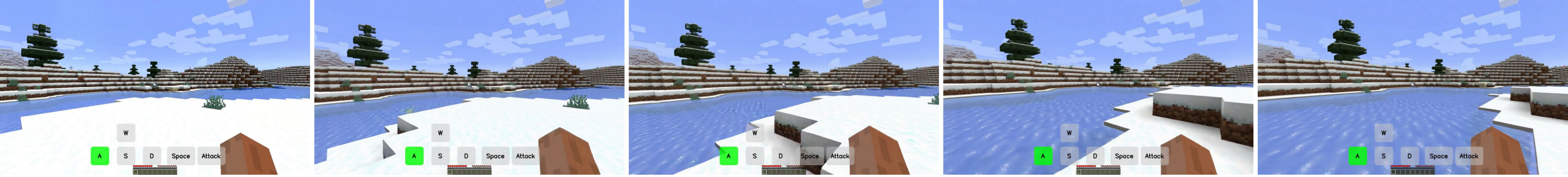}
    \caption{Left}
    \vspace{0.1in}
\end{subfigure}  

\begin{subfigure}{0.99\linewidth}
    \centering
    \includegraphics[width=\linewidth]{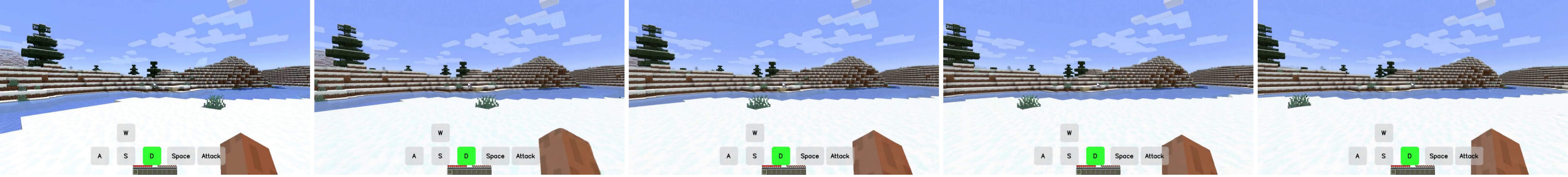}
    \caption{Right}
    \vspace{0.1in}
\end{subfigure}  

\begin{subfigure}{0.99\linewidth}
    \centering
    \includegraphics[width=\linewidth]{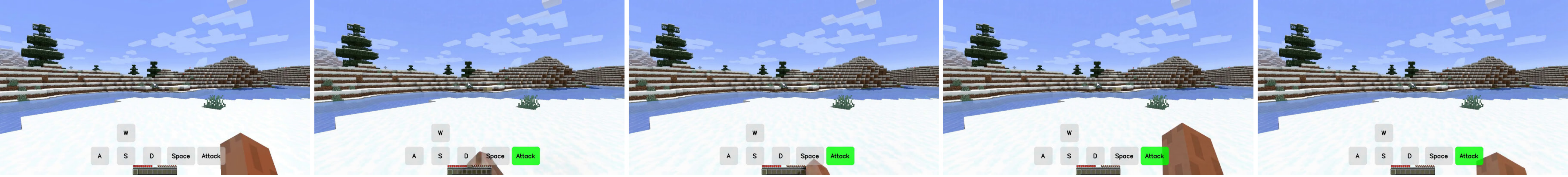}
    \caption{Attack}
    \vspace{0.1in}
\end{subfigure}  

\begin{subfigure}{0.99\linewidth}
    \centering
    \includegraphics[width=\linewidth]{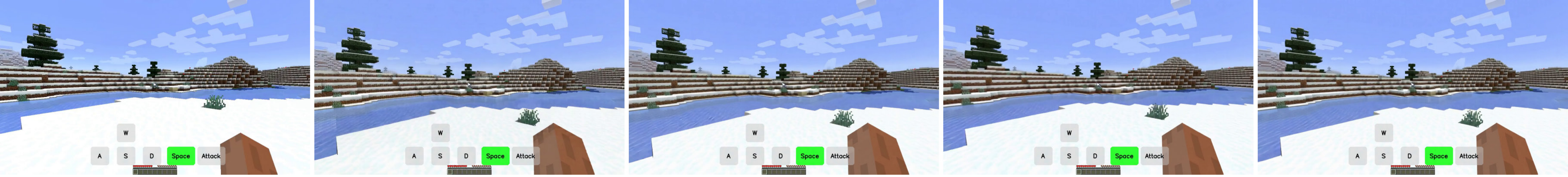}
    \caption{Jump} 
\end{subfigure}  
 
\caption{Demonstration of \textit{Matrix-Game}'s controllable video generation conditioned on various keyboard actions, including \textit{forward}, \textit{back}, \textit{left}, \textit{right}, \textit{jump}, and \textit{attack}. The model accurately responds to user's control signals and generates coherent motion patterns.}
\label{fig:demo_keyboard_actions}
\end{figure}

\begin{figure}[p]
\centering

\begin{subfigure}{0.99\linewidth}
    \centering
    \includegraphics[width=\linewidth]{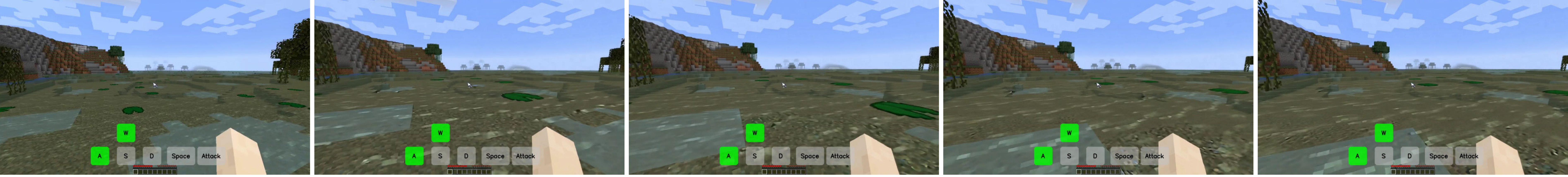}
    \caption{Forward + Left}
    \vspace{0.1in}
\end{subfigure}

\begin{subfigure}{0.99\linewidth}
    \centering
    \includegraphics[width=\linewidth]{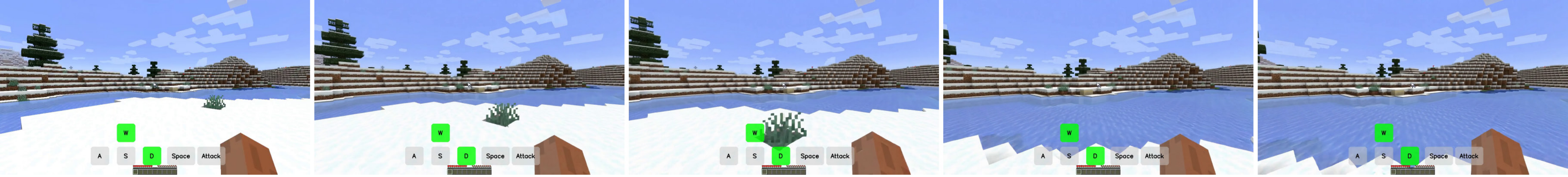}
    \caption{Forward + Right}
    \vspace{0.1in}
\end{subfigure}

\begin{subfigure}{0.99\linewidth}
    \centering
    \includegraphics[width=\linewidth]{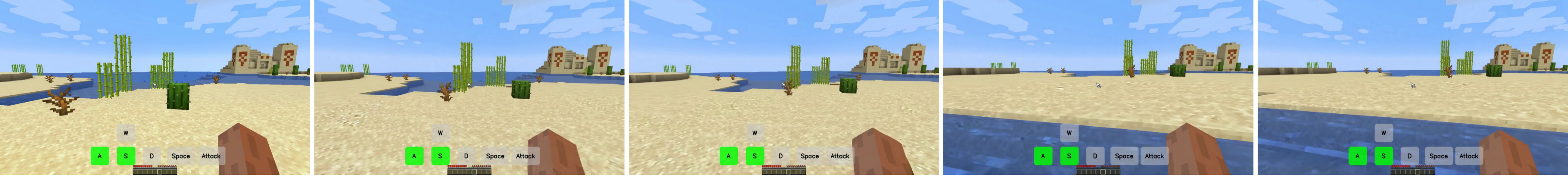}
    \caption{Backward + Left} 
\end{subfigure}

\begin{subfigure}{0.99\linewidth}
    \centering
    \includegraphics[width=\linewidth]{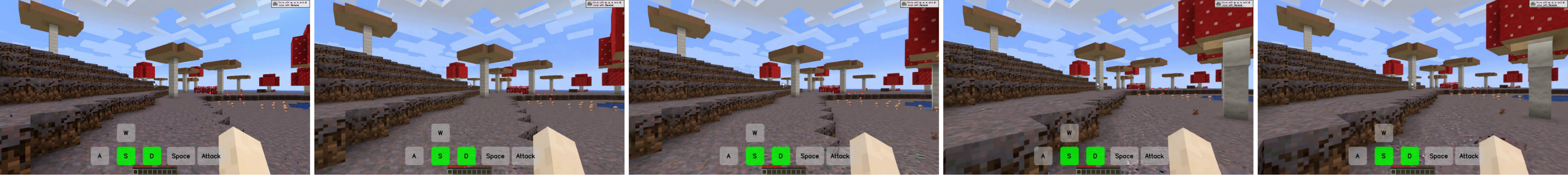}
    \caption{Backward + Right}
    \vspace{0.1in}
\end{subfigure}

\begin{subfigure}{0.99\linewidth}
    \centering
    \includegraphics[width=\linewidth]{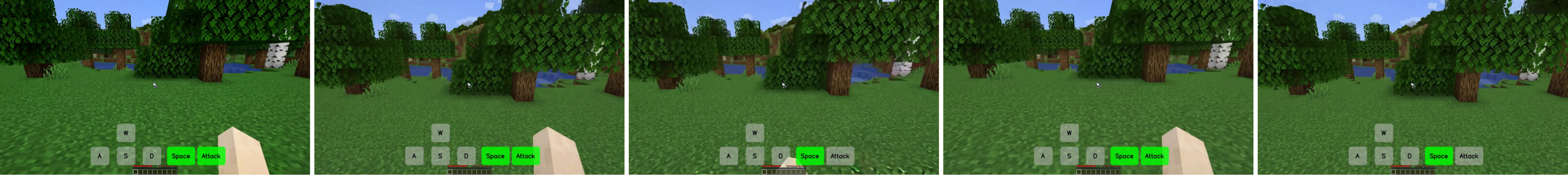}
    \caption{Jump + Attack}
    \vspace{0.1in}
\end{subfigure}

\begin{subfigure}{0.99\linewidth}
    \centering
    \includegraphics[width=\linewidth]{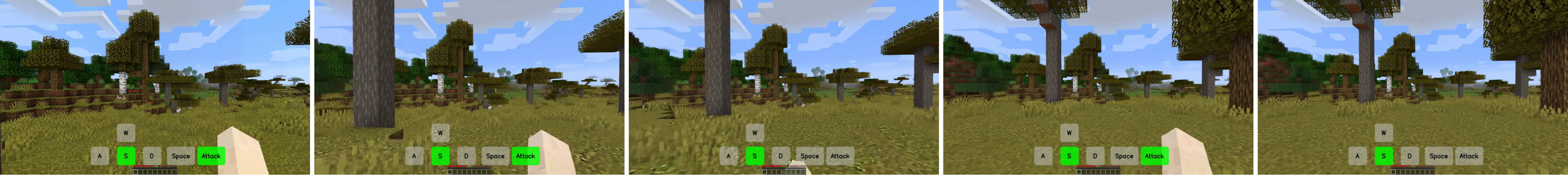}
    \caption{Backward + Attack}
    \vspace{0.1in}
\end{subfigure}

\begin{subfigure}{0.99\linewidth}
    \centering
    \includegraphics[width=\linewidth]{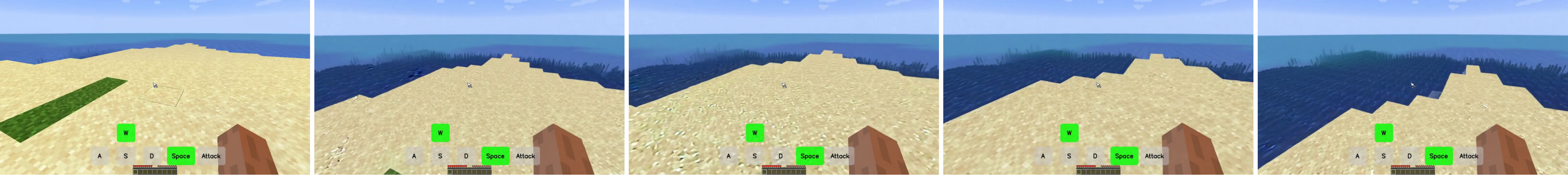}
    \caption{Forward + Jump}
    \vspace{0.1in}
\end{subfigure}

\begin{subfigure}{0.99\linewidth}
    \centering
    \includegraphics[width=\linewidth]{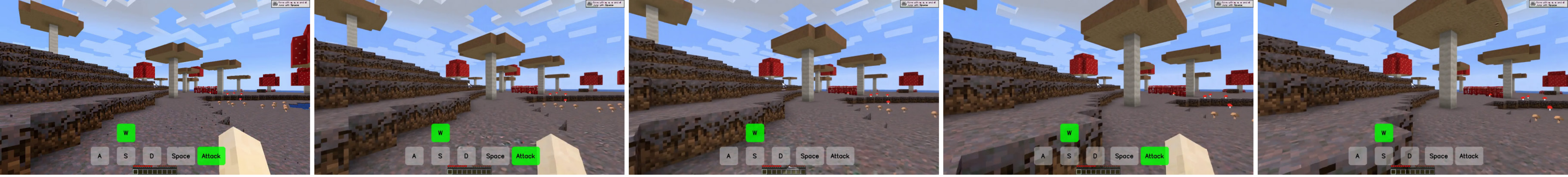}
    \caption{Forward + Attack}
    \vspace{0.1in}
\end{subfigure}

\caption{Demonstration of \textit{Matrix-Game}'s controllable video generation conditioned on complex actions, such as  \textit{forward + right}, \textit{backward + attack} and \textit{jump + attack},. The model effectively interprets complex user commands and generates coherent, action-consistent motion trajectories that reflect the intended behaviors.}

\label{fig:demo_mactions}
\end{figure}

To rigorously evaluate the action control capabilities, we compare {Matrix-Game} with OASIS~\cite{oasis2024} and MineWorld~\cite{guo2025mineworld} across two control modalities: keyboard actions (e.g., forward, backward, left, right, jump, and attack) and mouse-based camera movements (e.g., yaw and pitch angle shifts in 8 directions). As shown in Table~\ref{tab:control_signals}, {Matrix-Game} consistently achieves the highest control accuracy across all action types. Regarding keyboard control, our model attains over {88\%} accuracy on all actions, with particularly strong performance in fine-grained directions such as “forward” (99\%), “right” (96\%), and “jump” (95\%). For mouse-based camera movement, typically more challenging due to subtle pitch and yaw transitions, our model still maintains high accuracy, reaching {over 89\%} on all directions. These results highlight our model’s fine-grained controllability and strong alignment between action signals and generated behavior, making it well-suited for interactive world generation and agent conditioning in dynamic environments. For more qualitative results and video demonstrations, please refer to Figures~\ref{fig:demo_mouse_actions}-\ref{fig:demo_mactions} and our project website at \url{https://matrix-game-homepage.github.io}.

\begin{figure}[t]
\centering
\begin{subfigure}{0.245\linewidth}
    \centering
    \includegraphics[width=\linewidth]{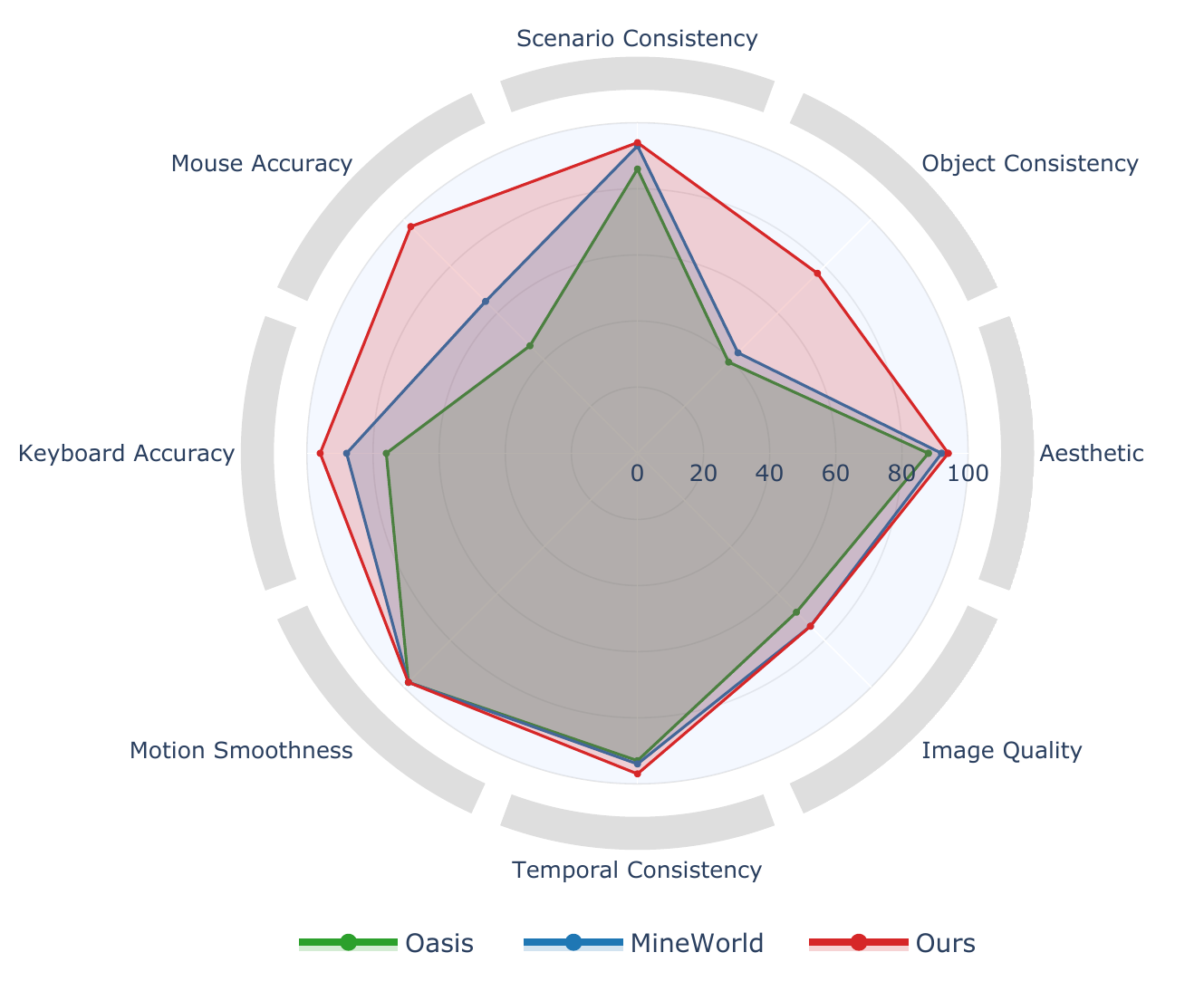}
    \caption{Desert}
\end{subfigure}
\begin{subfigure}{0.245\linewidth}
    \centering
    \includegraphics[width=\linewidth]{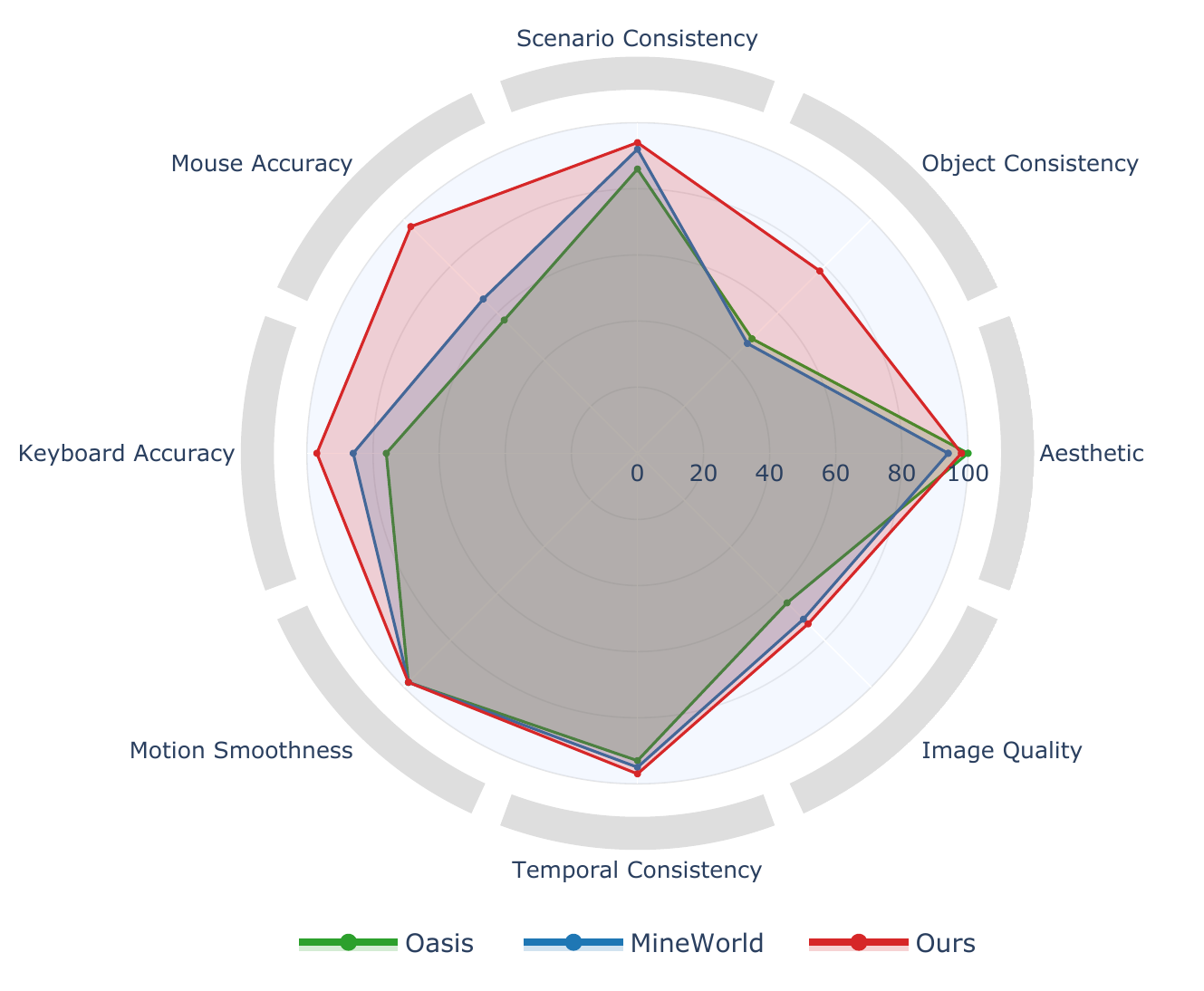}
    \caption{Beach}
\end{subfigure}
\begin{subfigure}{0.245\linewidth}
    \centering
    \includegraphics[width=\linewidth]{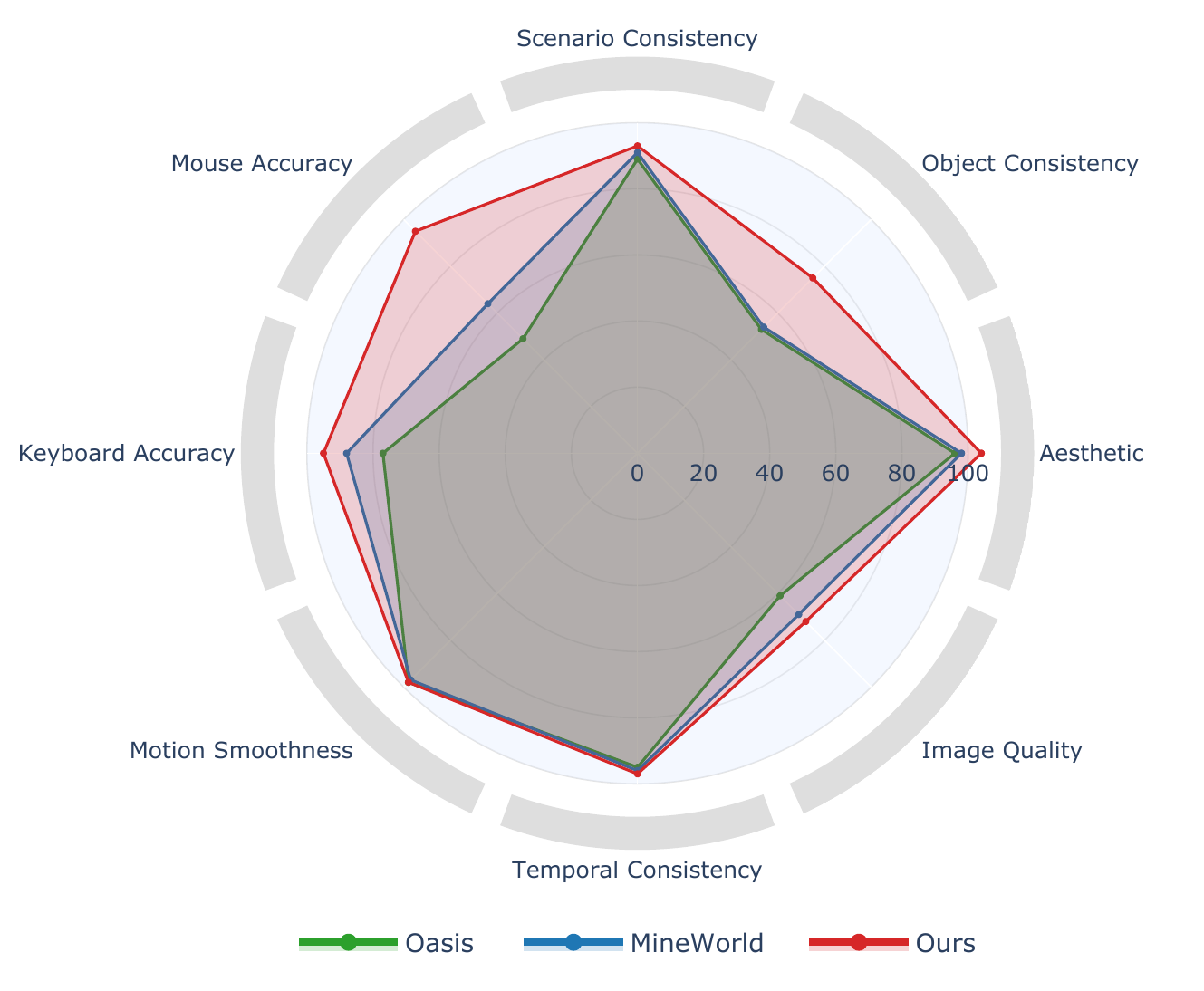}
    \caption{Forest}
\end{subfigure}
\begin{subfigure}{0.245\linewidth}
    \centering
    \includegraphics[width=\linewidth]{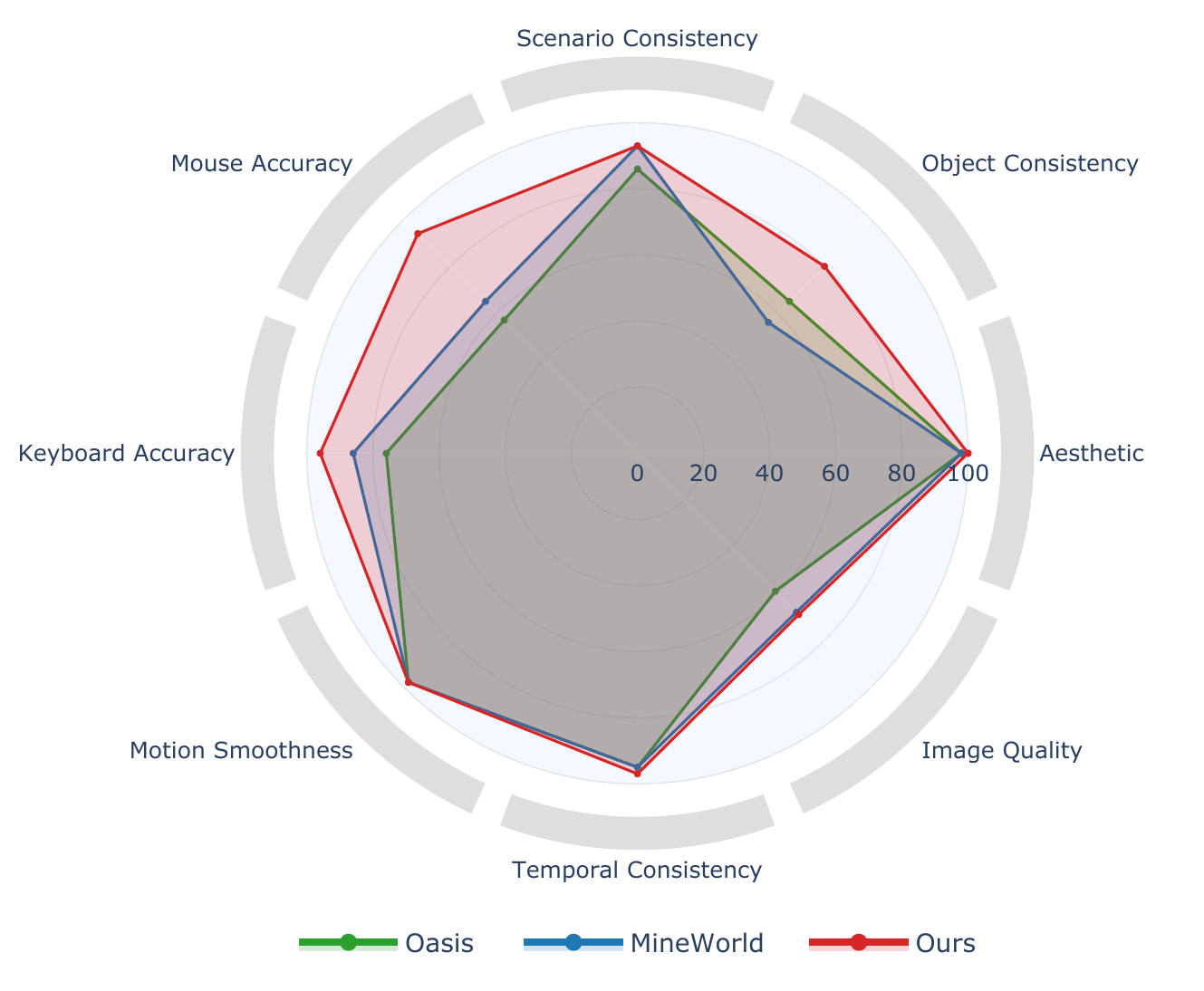}
    \caption{Hills}
\end{subfigure}

\begin{subfigure}{0.245\linewidth}
    \centering
    \includegraphics[width=\linewidth]{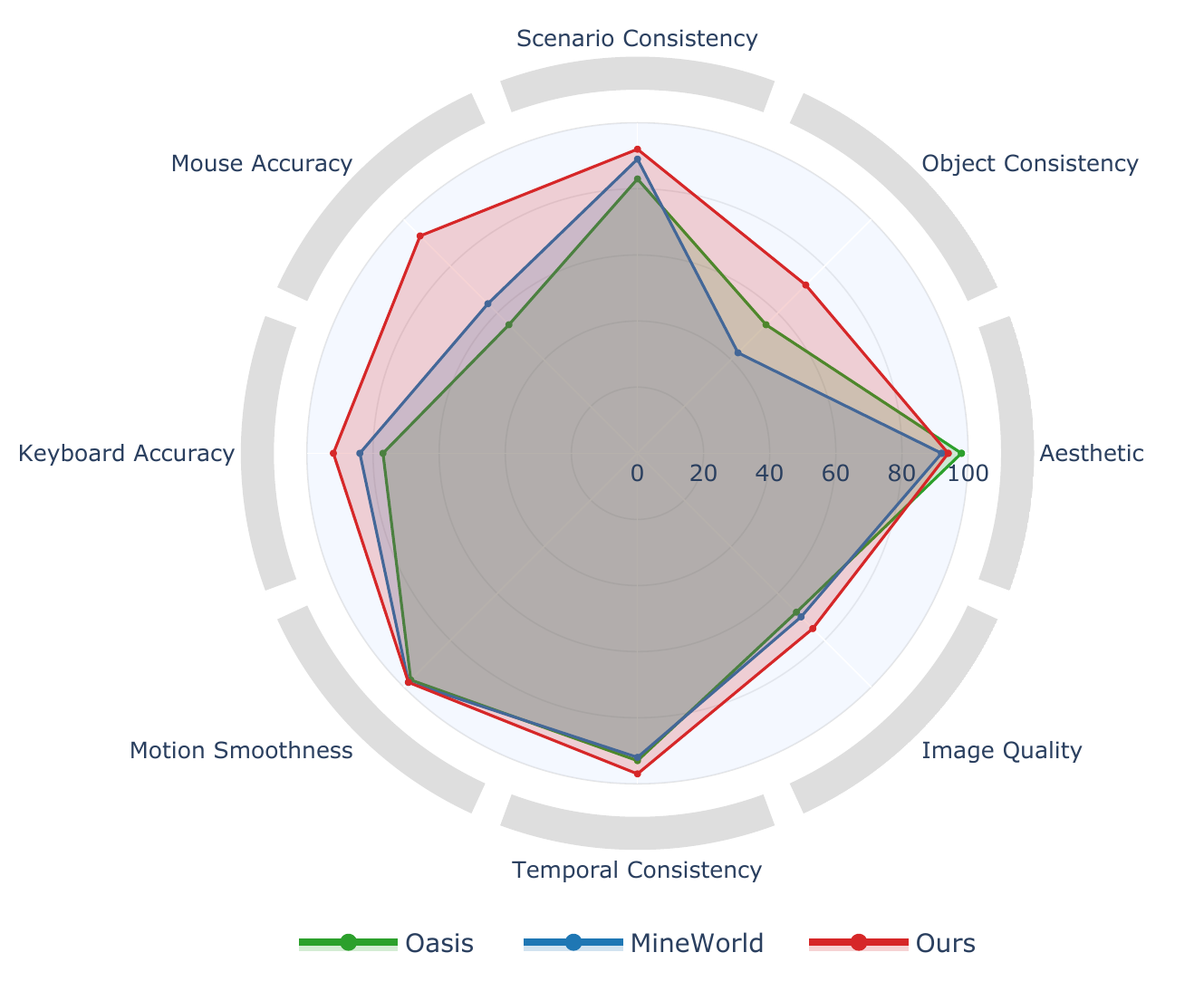}
    \caption{Icy}
\end{subfigure}
\begin{subfigure}{0.245\linewidth}
    \centering
    \includegraphics[width=\linewidth]{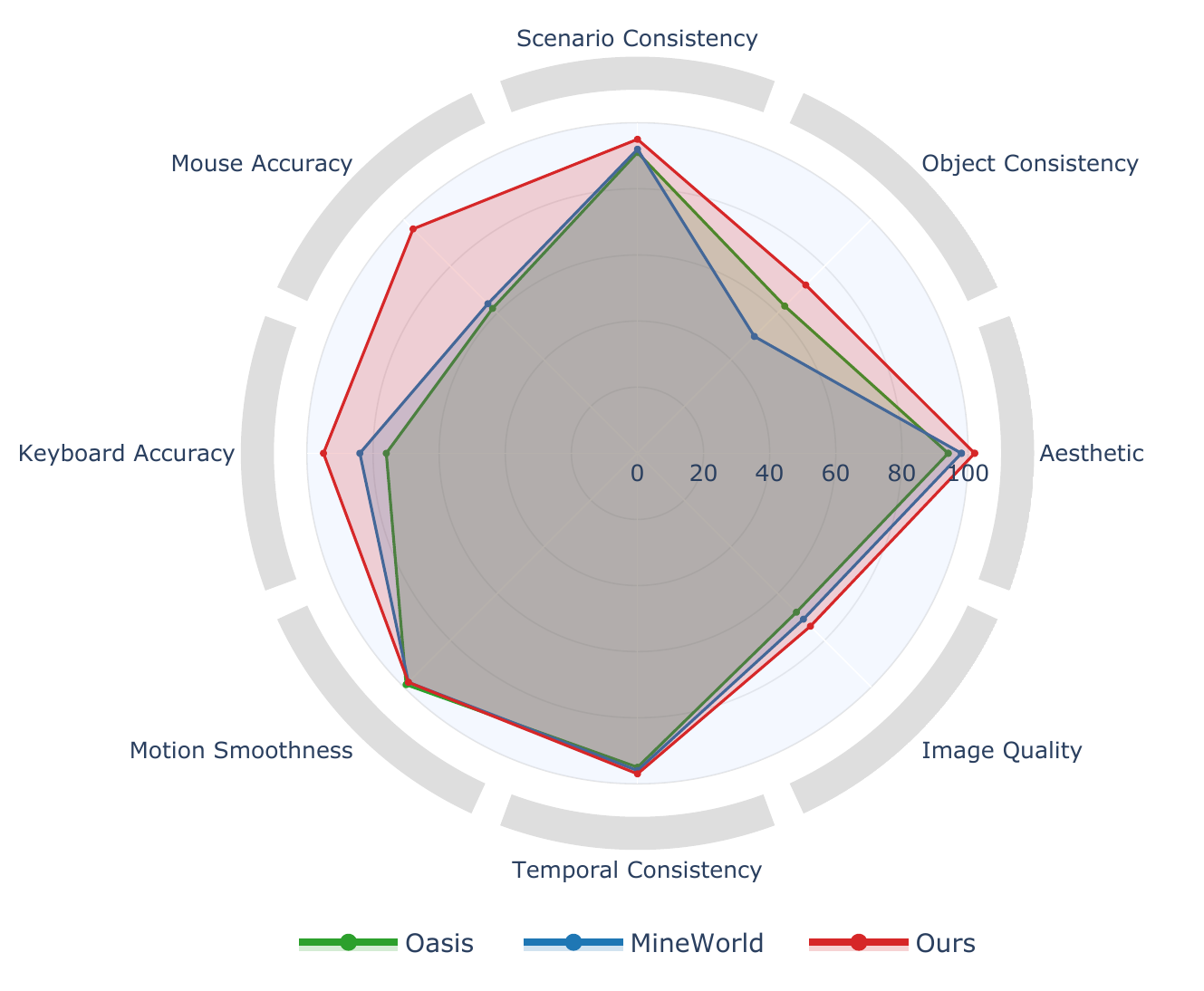}
    \caption{Plain}
\end{subfigure}
\begin{subfigure}{0.245\linewidth}
    \centering
    \includegraphics[width=\linewidth]{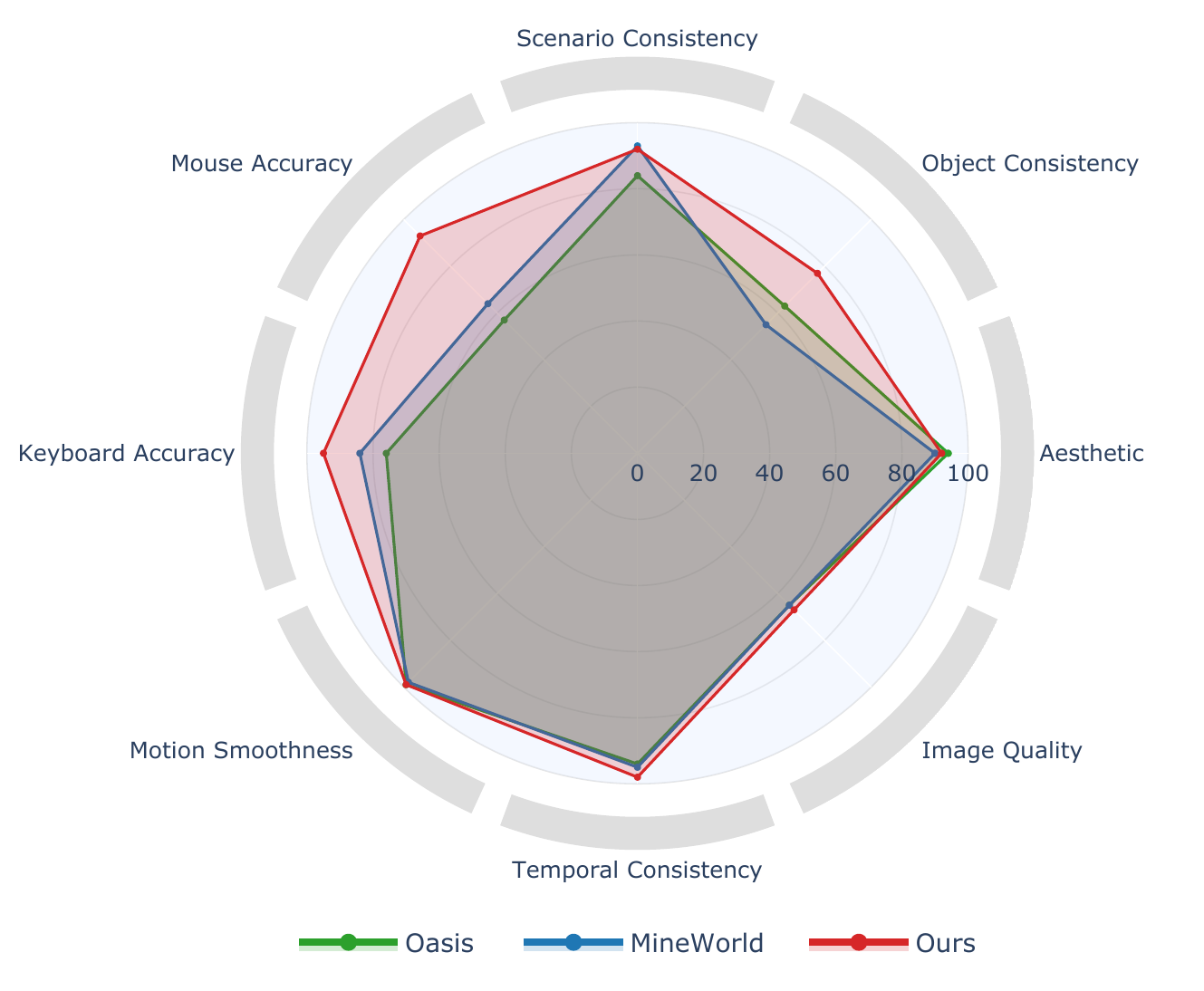}
    \caption{River}
\end{subfigure}
\begin{subfigure}{0.245\linewidth}
    \centering
    \includegraphics[width=\linewidth]{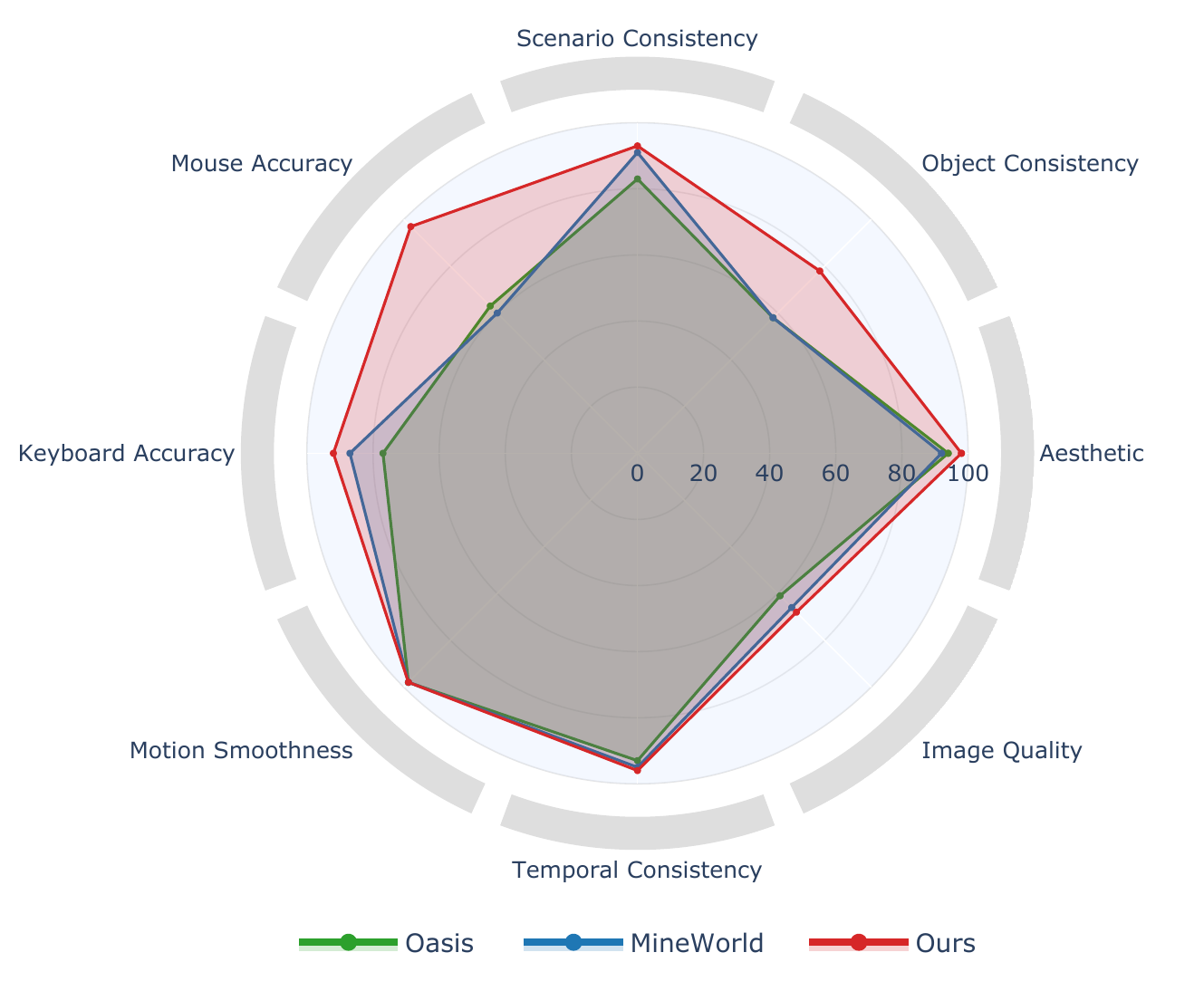}
    \caption{Mushroom}
\end{subfigure}

\caption{{GameWorld Score} across eight scenarios. Each radar chart shows performance over 8 axes: Image Quality, Aesthetic (scaled $\times$2 for clarity), Temporal Consistency, Motion Smoothness, Keyboard/Mouse Accuracy, and Object/Scenario Consistency. Our method consistently shows superior controllability and physical consistency while preserving high visual and temporal quality.}
\label{fig:gameworld_radar_all}
\end{figure}

\subsection{Scenario Generalization}
To assess the generalization capability of our model across diverse game environments, we evaluate it on eight distinct Minecraft scenarios: \textit{beach}, \textit{desert}, \textit{forest}, \textit{hills}, \textit{icy}, \textit{mushroom}, \textit{plains}, and \textit{river}. These scenarios cover a wide range of terrain types, object distributions, and interaction dynamics, providing a comprehensive testbed for evaluating world generation performance.
As shown in Figure~\ref{fig:gameworld_radar_all}, our model consistently outperforms existing open-source baselines, OASIS~\cite{oasis2024} and MineWorld~\cite{guo2025mineworld}, across all eight scenarios.

Notably, {Matrix-Game} achieves substantial improvements in {action controllability}, measured via keyboard and mouse accuracy, and in {physical consistency}, which assesses the model’s ability to maintain object and scenario consistency over time. These gains underscore the model’s robustness in responding accurately to diverse and fine-grained control signals while preserving geometric coherence across a wide range of world scenarios.
Beyond these core capabilities, our model also demonstrates strong performance in {visual quality}, achieving higher scores in both image fidelity and aesthetic appeal. This ensures that generated frames are not only semantically consistent but also visually compelling. Furthermore, the model maintains superior {temporal smoothness}, producing videos that transition fluidly between frames with minimal flicker or abrupt motion artifacts.
Together, these results highlight the comprehensive generalization ability of {Matrix-Game}, making it well-suited for complex and dynamic tasks in interactive world generation. 

\begin{figure}[p]
\centering
\begin{subfigure}{0.99\linewidth}
    \centering
    \includegraphics[width=\linewidth]{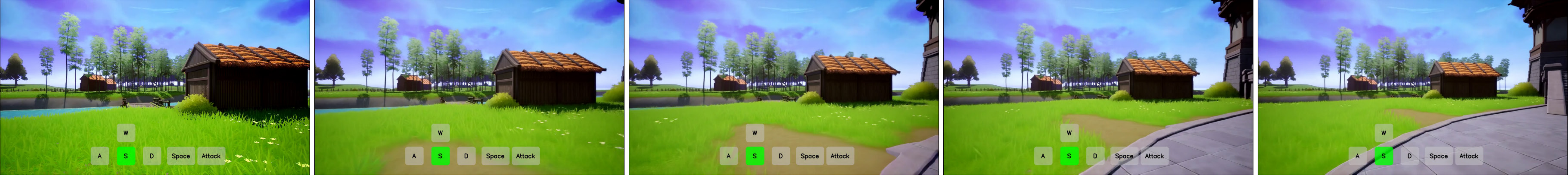}
    \caption{Village}
    \vspace{0.1in}
\end{subfigure}

\begin{subfigure}{0.99\linewidth}
    \centering
    \includegraphics[width=\linewidth]{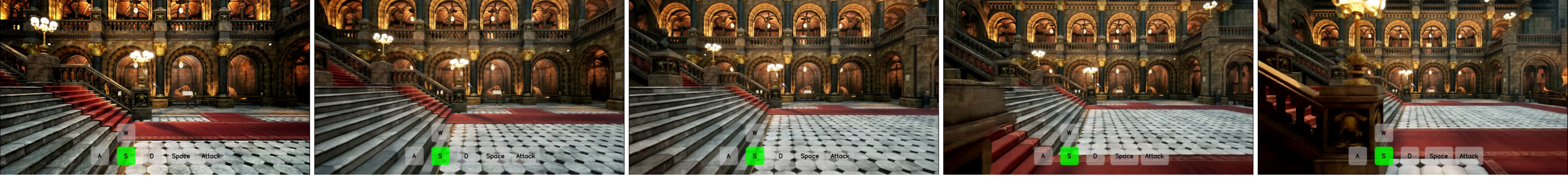}
    \caption{Hall}
    \vspace{0.1in}
\end{subfigure}  

\begin{subfigure}{0.99\linewidth}
    \centering
    \includegraphics[width=\linewidth]{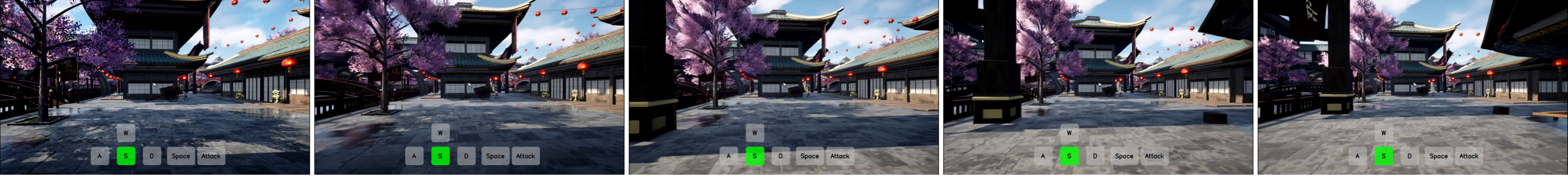}
    \caption{Oriental Palace}
    \vspace{0.1in}
\end{subfigure}  

\begin{subfigure}{0.99\linewidth}
    \centering
    \includegraphics[width=\linewidth]{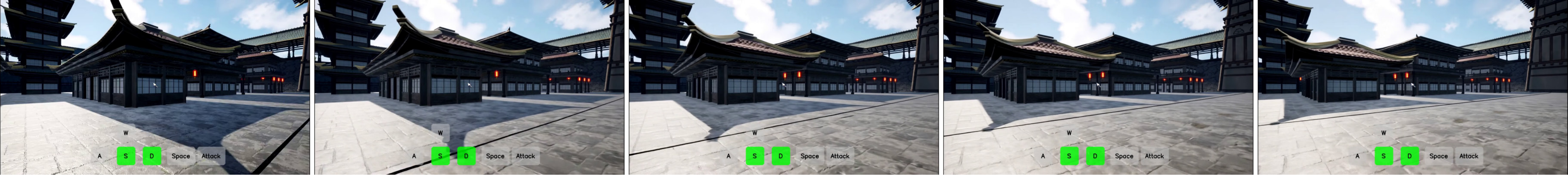}
    \caption{Oriental Palace}
    \vspace{0.1in}
\end{subfigure}  

\begin{subfigure}{0.99\linewidth}
    \centering
    \includegraphics[width=\linewidth]{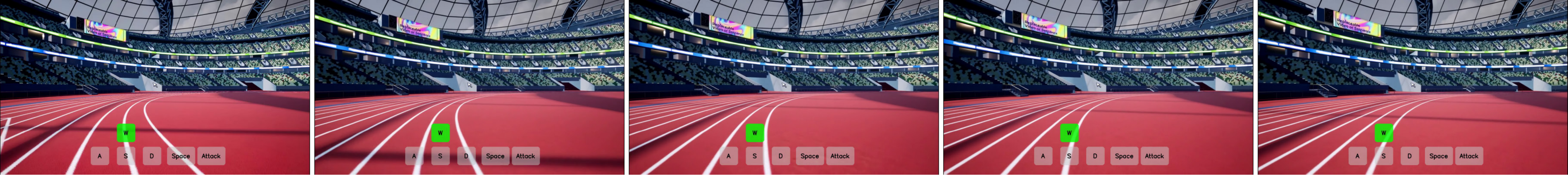}
    \caption{Stadium}
    \vspace{0.1in}
\end{subfigure}  
 
\begin{subfigure}{0.99\linewidth}
    \centering
    \includegraphics[width=\linewidth]{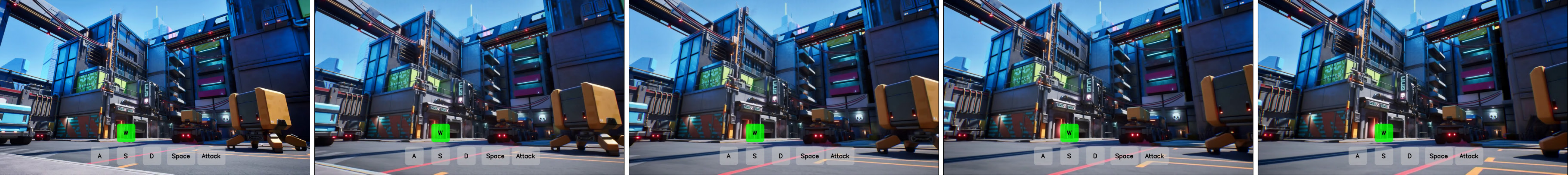}
    \caption{City}
    \vspace{0.1in}
\end{subfigure}  

\begin{subfigure}{0.99\linewidth}
    \centering
    \includegraphics[width=\linewidth]{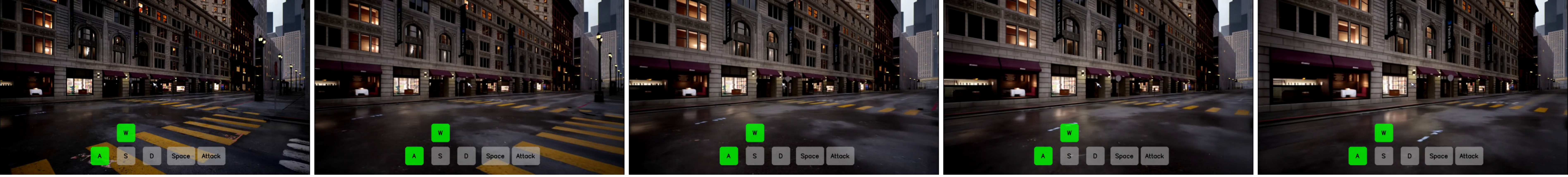}
    \caption{City}
    \vspace{0.1in}
\end{subfigure}  

\begin{subfigure}{0.99\linewidth}
    \centering
    \includegraphics[width=\linewidth]{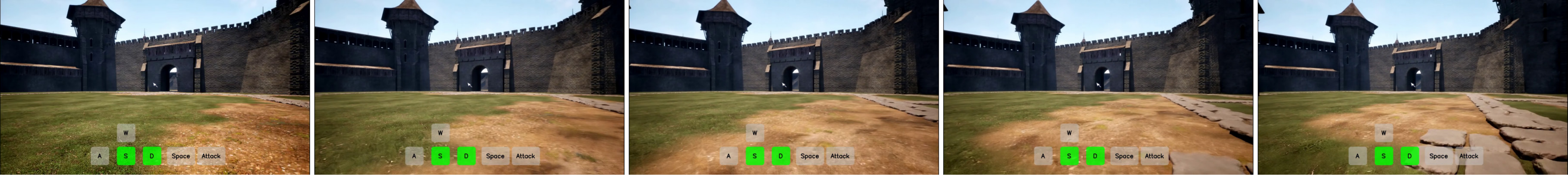}
    \caption{Ancient City Wall}
    \vspace{0.1in}
\end{subfigure}  
\caption{Controllable world generation results of \textit{Matrix-Game} across distinct Unreal scenarios. These demos illustrate the model's ability to handle diverse environments, ranging from \textit{Village}, \textit{City}, and \textit{Stadium} to \textit{Oriental Palace} and \textit{Hall}.}
\label{fig:demo_unreal}
\end{figure}

For more qualitative demonstrations across various game scenarios, please refer to Figures~\ref{fig:demo_scenarios},~\ref{fig:demo_unreal} and visit our project website at \url{https://matrix-game-homepage.github.io}.

\begin{figure}[t]
    \vspace{-0.2in}
\centering
\begin{subfigure}{0.99\linewidth}
    \centering
    \includegraphics[width=\linewidth]{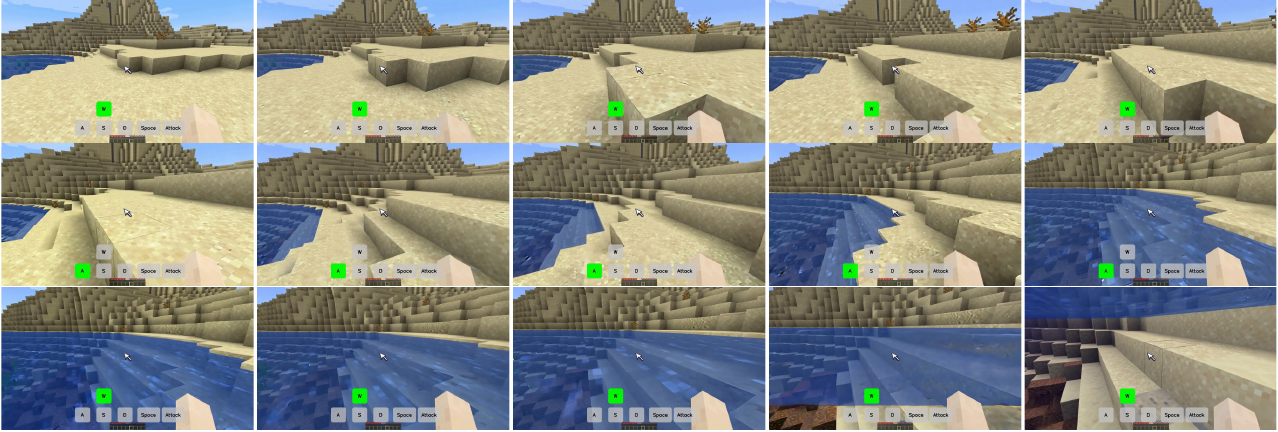}
    \caption{Keyboard actions: forward $\rightarrow$ left $\rightarrow$ forward}
    \vspace{0.1in}
\end{subfigure} 

\begin{subfigure}{0.99\linewidth}
    \centering
    \includegraphics[width=\linewidth]{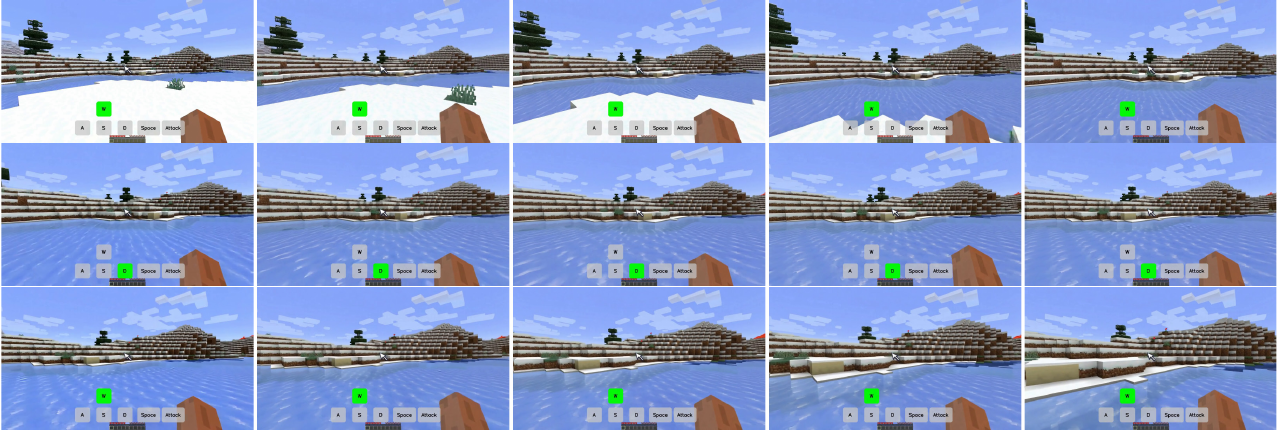}
    \caption{Keyboard actions: forward $\rightarrow$ right $\rightarrow$ forward}
    \vspace{0.1in}
\end{subfigure} \begin{subfigure}{0.99\linewidth}
    \centering
    \includegraphics[width=\linewidth]{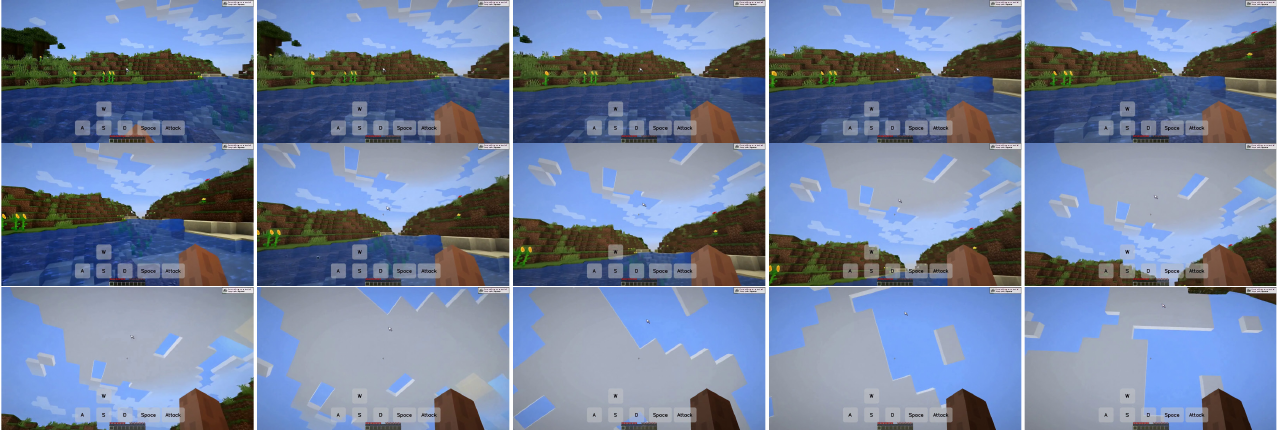}
    \caption{Mouse movement actions: (camera $\rightarrow$) $\rightarrow$ (camera $\uparrow$) $\rightarrow$ (camera $\nearrow$)}
\end{subfigure}  
\caption{Auto-regressive generation results of \textit{Matrix-Game} across three action-conditioned segments. Each segment is generated independently based on the preceding motion context and current action signal. Despite the segment-wise generation, \textit{Matrix-Game} maintains strong temporal consistency between segments and faithfully follows user-specified control across long sequences.}
    \vspace{-0.1in}
\label{fig:demo_ar}
\end{figure}

\subsection{Autoregressive Generation for Long Videos}
To support long-form video generation, we design our model to operate in an {autoregressive multi-segment} setting, where the video is generated in consecutive segments conditioned on previously synthesized frames. As shown in Figure~\ref{fig:demo_ar}, our approach maintains  strong {local temporal consistency} between segments, effectively bridging the boundaries between consecutive chunks. This results in smooth and visually coherent long-range video outputs without geometry misalignment or abrupt motion shifts, thus can  sustain extended interactions and dynamic control behaviors across time.

Please refer to the video demos of long-form video generation at our project website: \url{https://matrix-game-homepage.github.io}.

\begin{figure}[t]
\centering
\begin{subfigure}{0.99\linewidth}
    \centering
    \includegraphics[width=\linewidth]{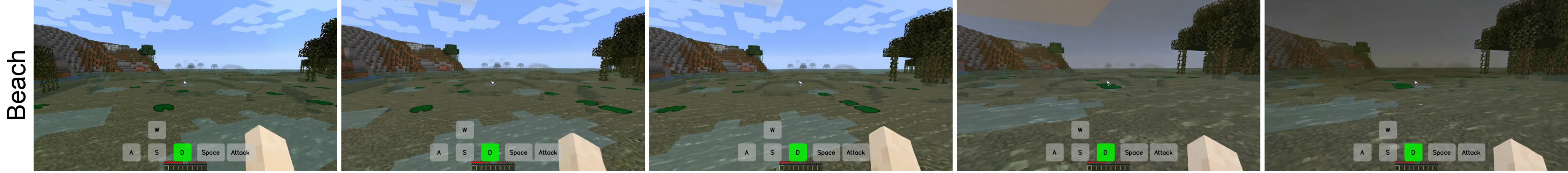}
    \caption{Edge case generalization}
    \vspace{0.1in}
\end{subfigure}
\begin{subfigure}{0.99\linewidth}
    \centering
    \includegraphics[width=\linewidth]{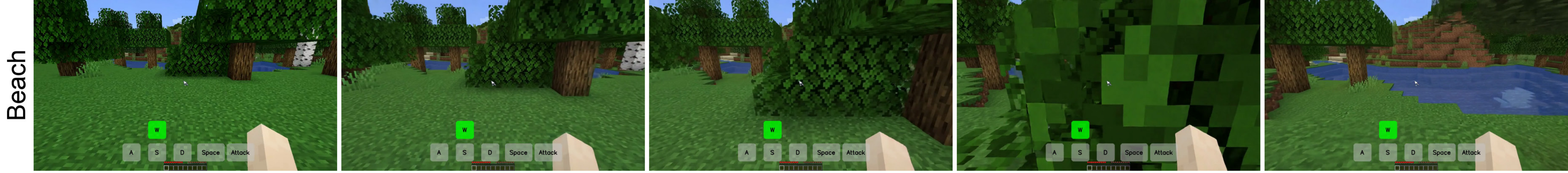}
    \caption{Physics understanding}
\end{subfigure}  
\caption{Failure cases of Matrix-Game.  (a) \textit{Edge case}: the model may fail to maintain temporal consistency in underrepresented or unfamiliar scenarios. (b) \textit{Physics understanding}: the agent walks through leaves, indicating that there is room to improve the modeling of physical interaction.}
\label{fig:failure_cases}
\end{figure}

\subsection{Discussions on Failure Cases}\label{failure_cases}

\textbf{Edge case generalization.}
While {Matrix-Game} demonstrates strong generalization across a wide range of Minecraft scenarios, we do observe certain failure cases in visually complex or underrepresented scenarios. As shown in  Figure~\ref{fig:failure_cases}(a), the model may occasionally struggle with precise controllability or spatial consistency in rare biomes or edge cases, typically due to insufficient data coverage. 
We believe these are natural challenges arising from the open-ended and procedurally generated nature of Minecraft. Although our dataset construction emphasizes diversity and balance, it is inherently difficult to cover all possible world variations. To improve generalization in such cases, we plan to augment our training datasets by scaling up the unlabeled gameplay video clips and creating more controllable data across diverse Minecraft scenes. We also envision adopting a continual training strategy to gradually adapt the model to newly encountered scenarios over time.

\textbf{Physics understanding.}
Although our model performs well in terms of object and consistency consistency, there is still room to further enhance its understanding of physical dynamics, particularly in interactions such as object collisions or terrain traversal.  For example, as shown in Figure~\ref{fig:failure_cases}(b), we observe cases where the generated agent walks through leaves. These issues are partly due to the scarcity of high-fidelity, physics-supervised data in the current dataset. We believe these aspects can be further improved with richer, physics-aware training data and more explicit modeling of environment constraints. In future work, we plan to curate more physics-aware scenarios, particularly those involving object collisions, support, and material interactions, and extend the {GameWorld Score} benchmark to include targeted assessments of physical rule understanding. We see this as an important next step toward building models that are not only controllable and coherent, but also physically grounded.

\section{Conclusions and Future Work}

In this work, we introduce Matrix-Game, a novel world foundation model tailored for interactive video generation in open-ended game environments. Alongside the model, we construct the Matrix-Game-MC dataset, a large-scale and richly annotated corpus designed to support action-controllable generation in Minecraft-style environments. To facilitate standardized evaluation in this emerging field, we also develop GameWorld Score, a comprehensive benchmark that captures key aspects of perceptual quality, temporal coherence, controllability, and physical consistency. We will release both the model weights and benchmark toolkit to the community, with the goal of advancing future research in interactive world generation.

\paragraph{Future Work.} While our method demonstrates strong results across a wide range of scenarios, it is not without limitations. As discussed in Section~\ref{failure_cases}, our model occasionally faces challenges in generalization and physical rule understanding, particularly in visually rare or structurally complex environments, largely due to limited training coverage. Addressing these limitations will require expanded data collection, targeted scenario enrichment, and continual training. Beyond this, we highlight three promising directions for future work below.

\textit{Long-term temporal consistency}. Maintaining coherence over extended video sequences remains a fundamental challenge for interactive world generation. While some recent work~\cite{xiao2025worldmem,gu2025long} has attempted to address this issue, current solutions remain for further exploration. We plan to enhance our model architecture by incorporating longer motion contexts or designing memory-based mechanisms to improve consistency across long-range temporal consistency.

\textit{Action space enrichment}. Our current approach supports six types of keyboard actions and a mouse control space with a limited range of directional values. However, real-world game environments, especially in Minecraft, feature a richer and more nuanced mouse spectrum. We aim to expand our keyboard action  space and enable mouse control with a broader and more continuous value range, thereby improving the precision and expressiveness of user interactions.

\textit{Beyond Minecraft}. While Minecraft provides a well-suited platform for scalable data construction and grounded interaction, it remains a simplified sandbox world. To push the boundaries of controllable generation, we plan to extend our framework to more complex game environments such as Black Myth: Wukong, racing simulators, and multi-agent combat games like CS:GO. These platforms offer richer visual dynamics, complex action semantics, and multi-agent interactions, posing exciting new challenges and opportunities for next-generation world models.

We hope that our contributions, spanning model design, dataset construction, evaluation protocols, and open-source resources, can provide a solid foundation and inspire further progress toward general-purpose, controllable, and physically grounded world generation.

{ 
\small
\bibliography{neurips_2025}
\bibliographystyle{plain} 
}

\end{document}